\documentclass[journal]{IEEEtran}
\usepackage{cite}
\usepackage{times}
\usepackage{epsfig}
\usepackage{graphicx}
\usepackage{amsmath}
\usepackage{amssymb}
\usepackage{algorithm, algorithmic}
\usepackage{enumerate}
\usepackage{multirow}
\usepackage{xcolor}
\usepackage{graphics}
\usepackage{epsfig}
\usepackage{epstopdf}
\usepackage{subfigure}
\usepackage[switch]{lineno}
\usepackage{url}

\usepackage{caption}
\usepackage{colortbl}
\definecolor{mygray}{gray}{.9}
\definecolor{mygray1}{gray}{.8}

\makeatletter
\def\hlinew#1{%
  \noalign{\ifnum0=`}\fi\hrule \@height #1 \futurelet
   \reserved@a\@xhline}
\makeatother

\begin{document}

\title{Transductive Zero-Shot Hashing For Multilabel Image Retrieval}

\author{Qin~Zou,
        Ling~Cao,
        Zheng~Zhang,
        Long~Chen,
        Song~Wang\\
        $\ $ \\
        {\color{blue}\url{https://github.com/qinnzou/Zero-Shot-Hashing}}
\thanks{Q.~Zou, L.~Cao and Z. Zhang are with the School of Computer Science, Wuhan University, Wuhan 430072, P.R.~China (E-mails: qzou@whu.edu.cn; lingcao@whu.edu.cn; zhengzhang@whu.edu.cn).}
\thanks{L.~Chen is with the School of Data and Computer Science, Sun Yat-Sen University, Guangzhou 518001,
P.R.~China (E-mail: chenl46@mail.sysu.edu.cn).}
\thanks {S.~Wang is with the Department of Computer Science and Engineering,
University of South Carolina, Columbia, SC 29201 USA (E-mail: songwang@cec.sc.edu).}

}

\markboth{IEEE Transactions on Neural Networks and Learning Systems, 2020}
{Shell \MakeLowercase{\textit{et al.}}: }


\maketitle

\begin{abstract}
Hash coding has been widely used in the approximate nearest neighbor search for large-scale image retrieval. Given semantic annotations such as class labels and pairwise similarities of the training data, hashing methods can learn and generate effective and compact binary codes. While some newly introduced images may contain undefined semantic labels, which we call unseen images, zero-shot hashing (ZSH) techniques have been studied for retrieval. However, existing ZSH methods mainly focus on the retrieval of single-label images and cannot handle multilabel ones. In this article, for the first time, a novel transductive ZSH method is proposed for multilabel unseen image retrieval. In order to predict the labels of the unseen/target data, a visual-semantic bridge is built via instance-concept coherence ranking on the seen/source data. Then, pairwise similarity loss and focal quantization loss are constructed for training a hashing model using both the seen/source and unseen/target data. Extensive evaluations on three popular multilabel data sets demonstrate that the proposed hashing method achieves significantly better results than the comparison methods.
\end{abstract}


\IEEEpeerreviewmaketitle

\section{Introduction} \label{sec:intro}
Hashing methods can transform high dimensional data into compact binary codes while preserving the similarity between them. With high computing efficiency and low storage cost, hashing methods have been widely used for large-scale image retrieval. A number of hashing methods have been proposed in the past decade~\cite{kulis2011kernelized,liu2012supervised,shen2013inductive,shen2015supervised}.

Existing hashing methods can be roughly divided into two categories: supervised~\cite{lai2015simultaneous,zhang2015bit,zou2019hash} and unsupervised~\cite{HWG11icml,datar2004locality,SHBC2015,weiss2009spectralNips,DH2015}. The supervised hashing methods incorporate human-annotated information, \textit{e.g.}, semantic labels and pairwise similarities, into the learning process to find an optimal hash function,
while the unsupervised methods often learn hash functions by exploiting the intrinsic manifold structure of the unlabeled data.
Generally, supervised methods can obtain much higher performance than the unsupervised ones.

\begin{figure}[!t]
  \centering
  \includegraphics[width=0.9\linewidth]{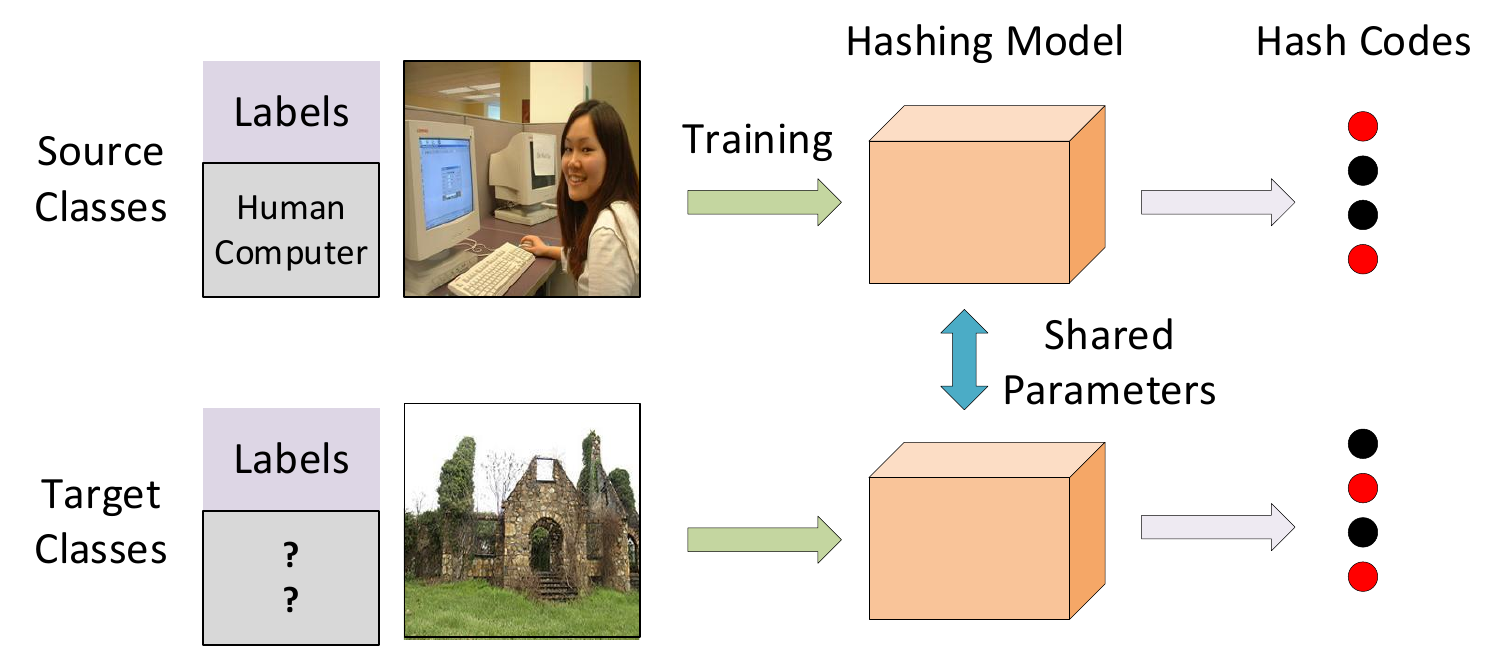}\\
  \caption{An illustration of the transductive zero-shot hashing. In the learning procedure, both the source and target data are used for training the hashing model. The categories (labels) of the target data are unknown to the learning system.}
  \label{structure1}
\end{figure}

In recent years, inspired by the remarkable success of deep neural networks in a broad range of computer-vision applications such as image classification~\cite{krizhevsky2012imagenet,Tsung2014coco,chen2019deep}, object detection~\cite{sun2014deep,zhou2019separability,shao2019saliency,chen2019surround}, semantic segmentation~\cite{long2015fully,zou2018deepcrack}, many supervised hashing methods turn to use deep neural networks for hash-code learning~\cite{xia2014supervised,zhao2015deep,zhu2016deep,liu2016deep,cao2017hashnet,yan2020deep-pami}.
These deep hashing methods have greatly advanced the retrieval performance on several popular benchmark datasets.

However, with the rapid emerging of new goods and new activities, images may contain concepts (or semantic labels) that are undefined before. For example, various commercial products with different shapes and appearances are released to the market every day, new sports with novel playing scenes are invented from time to time over the world. The images containing these new products or playing scenes are `unseen' as compared to the `seen' images holding pre-defined labels, and are supposed to be annotated with new labels for training the supervised learners.
Consequently, the supervised hashing methods may face tremendous challenges due to the lack of timely and reliable annotation of ground-truth labels for the unseen images.

Zero-shot learning (ZSL)~\cite{lampert2009learning} is a technique that can potentially solve or alleviate this problem. Zero-shot learning bridges the semantic gap between `seen' and `unseen' categories by transferring supervised knowledge from other modalities or domains, \textit{e.g.}, class-attribute descriptors and word vectors. For instance, the word embeddings of similar words that locate closely in the embedding space can capture the distributional similarity in the textual domain~\cite{pennington2014glove} based on a large-scale text corpus such as Wikipedia. Thus, such knowledge transfer can be used to capture the relationship between seen and unseen concepts, and can be helpful to handle unseen images in supervised learning.

For image retrieval under the circumstance of unseen images, some zero-shot hashing (ZSH) methods~\cite{yang2016zero,guo2017sitnet,pachori2018hashing,lai2018transductive,shao2019prl} have also been proposed. Nevertheless, these methods focus on single-label image retrieval, in which a one-to-one visual-semantic representation pair can satisfy the training of a hashing model. While in more complicated scenarios, an image often contains multiple object classes, and hence more complex semantics and their relationships.
How to represent the complex visual-semantics relationships for multi-label images, in a unified framework, is a difficult problem. To the best of our knowledge, there does not exist any work on multi-label zero-shot hashing.

Another important but easily ignored problem is that, since the underlying distributions of the source data and the target data are different,  learning a hash function from a naive knowledge transfer on the source domain without making it adaptive to the target domain may lead to severe domain-shift problems. The knowledge transfer across different domains has become a very important issue in many computer-vision problems~\cite{yan2019cross-tmm,yan2018fast-tmm,yan2020tmm-}, which also needs to be investigated in zero-shot hashing.

Considering the problems discussed above, we propose a novel transductive zero-shot hashing method (T-MLZSH) for multi-label image retrieval. Both the labeled source and the unlabeled target data are used in the training phase, as illustrated in Fig.~\ref{structure1}. The labeled source data are used to learn the relationship between visual images and semantic embeddings, and the unlabeled data of target classes are used to alleviate the domain-shift problem. More specifically, we first study a visual-semantic bridge via instance-concept coherence ranking on the source data. In instance-concept coherence ranking, a relatedness score for an image instance and a semantic concept is calculated for each image in the source data, where the score of an instance with a relevant label is larger than that of the same instance with an irrelevant label. Then, we can generate predicted labels for target data, and use these predicted labels as supervised information to guide the learning of hashing models. Moreover, we propose a focal quantization loss for fast and efficient hashing learning.

The contributions of this work lie in three-fold:
\begin{itemize}
  \item A transductive zero-shot hashing method (T-MLZSH) is proposed to solve the domain-shift problem in multi-label image retrieval. To the best of our knowledge, it is the first work studying the zero-shot hashing for multi-label image retrieval.
  \item An instance-concept coherence ranking algorithm is proposed for visual-semantic mapping, which can be used to predict the labels for unseen target data and hence improve the performance of zero-shot deep hashing.
  \item The proposed method obtains very promising results on three popular multi-label datasets, which constructs the benchmark for zero-shot multi-label image retrieval and paves the road for new research in this field.
\end{itemize}

The rest of this paper is organized as follows.
Section~\ref{sec:relate} briefly reviews the related work.
Section~\ref{sec:method} describes the neural network architecture for zero-shot image retrieval.
Section~\ref{sec:experiment} demonstrates the effectiveness of the proposed method by experiments. Finally,
Section~\ref{sec:conc} concludes the paper.

\begin{figure*}[!t]
	\centering
	\includegraphics[width=0.9\linewidth]{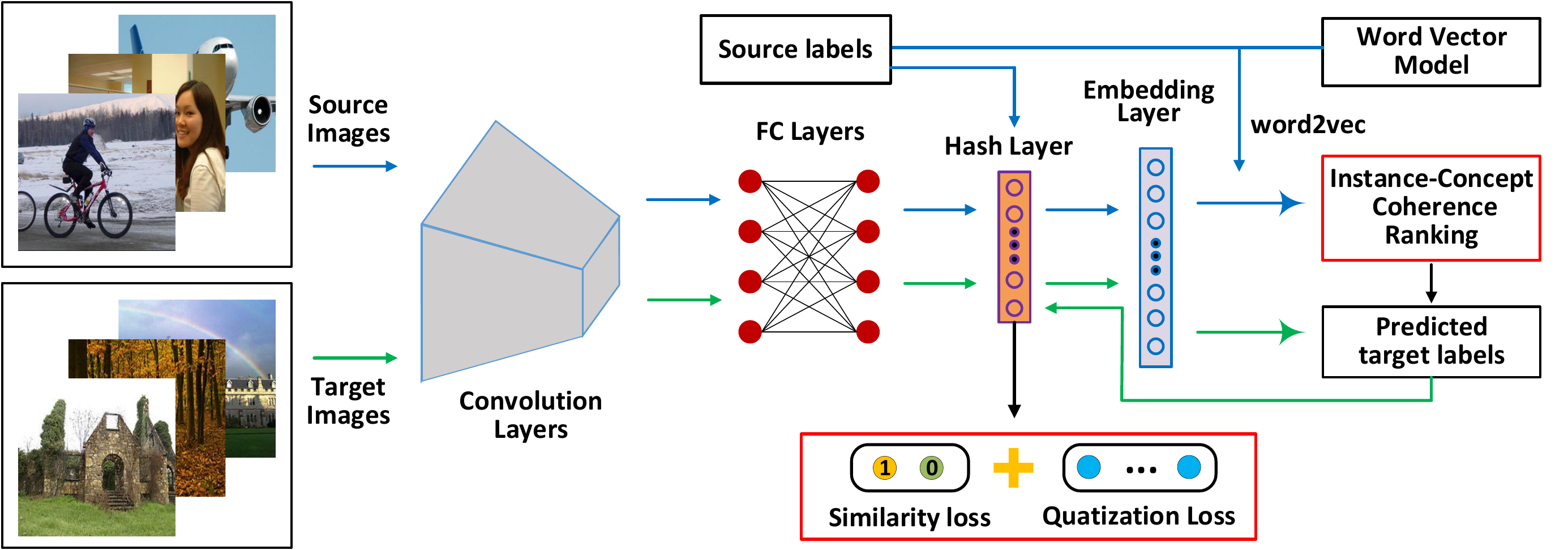}\\
	\caption{An overview of the proposed T-MLZSH network. At first, the model constructs a visual-semantic bridge via instance-concept coherence ranking on the source data. It calculates the relatedness scores between visual features and semantic word vectors, under the assumption that related instance-concept pairs should have larger relatedness scores. Given the learned instance-concept coherence relationship, the most relevant concepts for each image instance of the unlabeled target data can be selected to guide the similarity-preserving learning. The whole network can be trained in an end-to-end manner. }
	\label{structure}
\end{figure*}

\section{Related Work} \label{sec:relate}
\textbf{Zero-shot Learning.}
Zero-shot learning (ZSL), refers to training on the seen labeled data with seen labels, and testing on the unseen data with unseen labels, where there is no intersection between the seen label set and the unseen label set. The core of ZSL is to obtain the instance labels for the unseen data. According to the ways of obtaining the instances, ZSL can be divided into three categories~\cite{MiaoA,GongTransductive,Baroni2014Is}: projection-based methods~\cite{SutskeverExploiting,Frome2013DeViSE,Akata2014Evaluation}, instance-borrowing methods~\cite{Gan2016Recognizing, GaoZero, Yue2017Zero} and synthesizing methods~\cite{Akata2018Feature,zhu2018generative,Yue2017Synthesizing,ElgammalLearning}.

The projection-based methods obtain instances by projecting the elements in the feature space and semantic space into a common space. Generally, the feature space contains the instances of seen classes, and the semantic space contains the prototypes of seen/unseen classes. In~\cite{Frome2013DeViSE}, a hinge rank loss was constructed to learn a linear transformation from the feature space to the semantic space. In~\cite{Akata2014Evaluation}, a bilinear mapping function was learned by minimizing the loss of an SVM classifier. In~\cite{GuoProgressive}, the prediction noise and class bias of label embedding were decreased by deploying multiple classifiers in an ensemble manner. Instance-borrowing methods are based on the similarities of classes, in which the instances belonging to similar classes are taken as positive. In~\cite{Yi2016Recognizing}, instances in training were borrowed from the seen classes. The borrowed instances have high similarities with the unseen classes. Unlike instance-borrowing methods, the synthesizing methods create the pseudo instances for unseen classes, where the adversarial autoencoder and generative adversarial networks (GANs) are often used. In~\cite{MunjalSemantically}, an optimal latent space was learned to constrcut a bias-reducing generator network, which can reduce the hubness problem. In~\cite{ Mandal2019Out}, an out-of-distribution detector was introduced to reduce the effect of domain shift, and a GAN was employed to synthesize the unseen instances.

To handle multi-label images, the projection is much more complex for the projection-based ZSL. A possible solution is to follow the visual-semantic mapping strategy used in the single-label case. In~\cite{fu2015transductive}, the meaning of multiple labels for one instance was inferred by summing the word vector representations of individual labels. In~\cite{ren2015multi}, an alternative way was proposed to use the direct visual-semantic mapping, which first finds the corresponding area of each semantic label and then extracts the object-level visual presentation for visual-semantic mapping. Some other works try to utilize the co-occurrence among the labels. One typical work is the COSTA~\cite{mensink2014costa}, which constructs the linear projection matrix between the seen labels and unseen labels by statistic learning on the annotated datasets.

\textbf{Hashing-based image retrieval.}
Hashing methods for image retrieval can be roughly divided into two categories: the unsupervised and the supervised. The unsupervised hashing methods generate hash codes without any semantic labels. They use clustering techniques or projection strategies to transfer visual information to feature space and generate an optimized hash function to preserve the similarity in Hamming space~\cite{SHBC2015,LL2014,DH2015,SC2008,MS2012}. Some classical algorithms, such as SH~\cite{weiss2009spectralNips}, formulated hash encoding as a spectral graph-partitioning problem and learned a nonlinear mapping to preserve semantic similarity. Some other methods, e.g., SPQ~\cite{ning2017scalable} and muti-k-means~\cite{ercoli2017compact}, decomposed the feature space into a Cartesian product of low-dimensional subspaces and encoded high-dimensional feature vectors into binary codes by clustering-based subspace quantization~\cite{xia2020pami}. ECE~\cite{liu2015sequential} treated it as an optimization problem and combined the genetic programming with the boosting-based weight updating. DSTH~\cite{zhu2018exploring} advocated discrete hash codes and resorted to the semantics augment from auxiliary contextual modalities.

The supervised hashing methods use the annotation information to learn compact hash codes, which usually perform better than the unsupervised methods. Among these supervised methods, CNN based hashing methods have attracted more and more attention due to the powerful representation ability of deep neural networks~\cite{gui2016supervised}. According to the difference of input forms, the existing deep hashing methods can be divided into two kinds. One receives image triplets as the input of the network and generates hash codes by minimizing the triplet ranking loss~\cite{lai2015simultaneous,zhang2015bit}. These methods consume many computing resources and time to train the hashing model as there are enormous triplet combinations. The other receives the minibatch of images as input and uses the pairwise similarity between images as supervised information to learn the hash codes. Typical works of this form include the DHN~\cite{zhu2016deep}, DSH~\cite{liu2016deep}, and HashNet~\cite{cao2017hashnet}, etc. Considering that existing hashing methods often fall short in concentrating relevant images to be within a small Hamming ball, DCH~\cite{CVPR18DCH} built a pairwise cross-entropy loss on the Cauchy distribution to improve its capability on this point.

In recent years, hashing-based methods were also developed for multimodal retrieval, which transform high-dimensional data of different modalities into compact binary codes in a common Hamming space. The multimodal hashing supports image retrieval across other domains, e.g., text, video, audio, etc. Some representative work can be found in~\cite{7967838, 8423193, 9072435,9009571,wang15ijcai-,wang2016tip-}.

\textbf{Zero-shot hashing.}
To handle images with unseen categories, some deep learning-based methods formulate the hashing as an unsupervised problem~\cite{zhang2018unsupervised,shen2018unsupervised}. However, without using reliable supervised information, it is difficult to achieve satisfactory performance. Some other methods~\cite{yang2016zero,guo2017sitnet,Survey2010}, from a different perspective, consider it as a zero-shot hashing (ZSH) problem. The goal of ZSH~\cite{TMV2015,DHNCVPR17} is to transfer the model trained on the seen data to unseen data via other available knowledge, such as word vector representations. Since the underlying data distributions of the seen-categories and the unseen-categories are different, the hashing functions learned by the seen categories without any adaptation to the unseen categories may cause a domain-shift problem. To narrow the domain gap between seen data and unseen data, ZSH-DA~\cite{pachori2018hashing} first learns a zero-shot hashing model on seen data, and then learns the final hashing model with a domain-adaptation algorithm. In~\cite{lai2018transductive}, a transductive zero-shot hashing network (TZSH) was proposed, which contains a coarse-to-fine similarity mining to find the most presentative target examples of each unseen labels, and adds these presentative examples and its corresponding predicted labels to the process of supervised hashing learning.~\cite{shen2018zero} uses GCN to learn the zero-shot hashing model for sketch-image retrieval. Although the GCN method is very promising in exploring the relationship between semantic labels, it is not flexible and requires a pre-defined adjacency matrix of nodes and additional training costs.

\section{Transductive Multi-Label Zero-Shot Hashing}
\label{sec:method}
\subsection{Problem Definition}
Suppose $ \mathcal{D}^{s}$=$\{\mathcal{I}_{i}^{s}, \mathcal{Y}_{i}^{s} \}_{i=1}^{N_{s}} $ is a labeled source dataset including $N_s$ images, where $\mathcal{I}_{i}^{s} $ is an image and $\mathcal{Y}_{i}^{s}$ is the corresponding label annotated with one or more classes, and $ \mathcal{D}^{t}$=$\{\mathcal{I}_{i}^{t}\}_{i=1}^{N_{t}}$ is an unlabeled target dataset, which includes $N_t$ images of the unseen target classes and has the labels $\mathcal{Y}^{t}$ unknown.
In the zero-shot setting, the target and source classes are two mutually exclusive label sets, \textit{i.e.}, $\mathcal{\mathcal{Y}}^{t} \bigcap \mathcal{Y}^{s}$=$\phi $.
For hash-code learning, we construct the similarity matrix $\mathcal{S}$=$\{ s_{ij}| i,j$=$1,2,...,N_{s}$+$N_{t} \}$, where $s_{ij}$ = 1 denotes that the pairwise images $\mathcal{I}_{i}$ and $\mathcal{I}_{j}$ are similar, and $s_{ij}$ = 0 denotes they are dissimilar. The goal of T-MLZSH is to learn a mapping $\mathcal{F} : \mathcal{I} \mapsto \{-1, +1 \}^{M}$ on the labeled source dataset $ \mathcal{D}^{s}$ and the unlabeled dataset $ \mathcal{D}^{t}$ to encode an input image $\mathcal{I}_i$ into an $M$-bit binary code $\mathcal{F}(\mathcal{I}_i)$, with the pairwise similarity preserved.

Figure~\ref{structure} gives a flowchart of the proposed method.
The input images firstly go through the deep network with the stacked convolutional and fully-connected layers and are encoded as a high-dimensional feature representation. Then, the outputs of the last fully-connected layer are fed into a hashing layer for compact binary encoding. To transfer the knowledge from seen categories
to unseen categories and construct the bridge between visual and semantic modalities, we add a fully-connected layer after hashing layer, which maps features from hamming space to the common embedding space.

\subsection{Instance-Concept Coherence Ranking}
\begin{figure}[!t]
	\centering
	\includegraphics[width=0.99\linewidth]{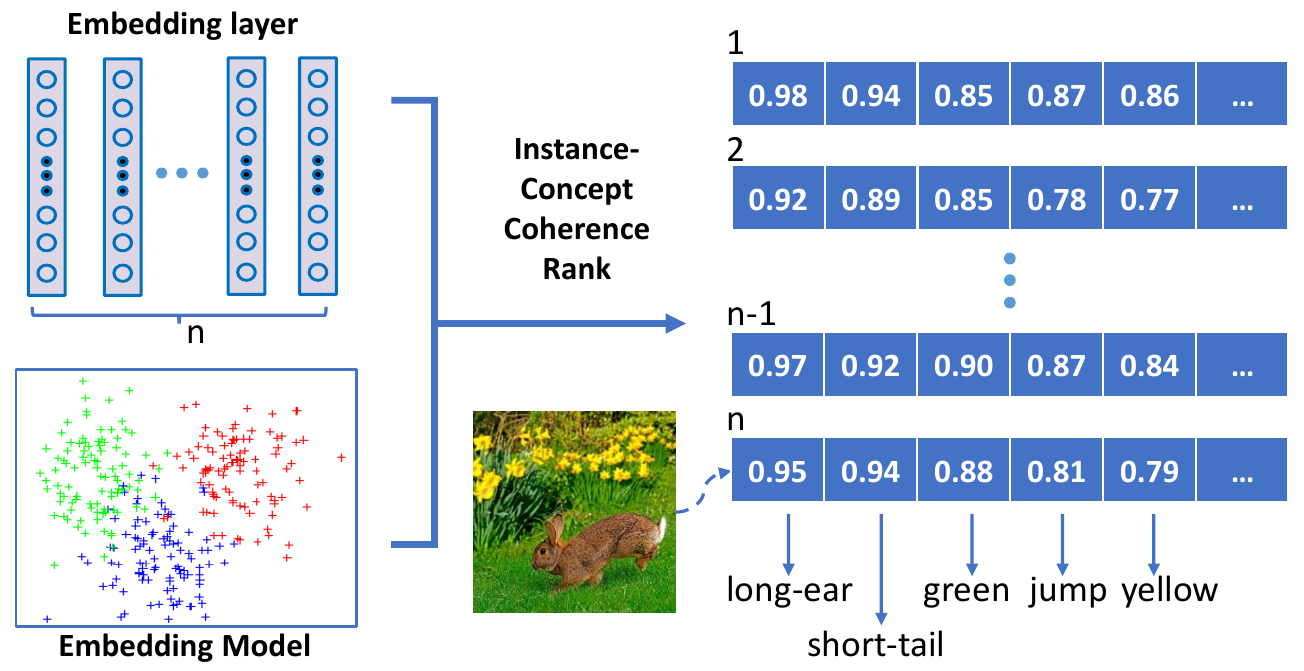}\\
	\caption{An illustration of the embedding model. It learns a mapping function where the scores with a relevant label should be higher than that with an irrelevant one. }
	\label{embedding}
\end{figure}

Since there are no label information for target images, we should firstly predict labels for these images by transferring the knowledge from the semantic representations to visual features, before learning a supervised hashing function. Let $v_{i}$ be the visual embedding of the $i$-th image instance $\mathcal{I}_{i}$ and $u_{j}$ be the semantic embedding of the $j$-th semantic concept, then we can calculate the relatedness score between $\mathcal{I}_{i}$ and the $j$-th semantic concept in the embedding space:
\begin{equation}\label{score}
  o_{ij} = < v_{i}, u_{j} >,
\end{equation}
where $\langle a, b \rangle = a^{T}b$ is the inner product operation. The semantic embeddings can be obtained from the existing word vector models, and the visual embeddings are variables that should be learned. During the training process, we can get a score list of source labels $\{o_{i1}, o_{i2}, \cdots , o_{iL_{s}}\}$, where $L_{s}$ is the number of seen labels. The goal of our embedding model is to learn a mapping function that scores with a relevant label should be higher than that with an irrelevant one, as illustrated by Fig.~\ref{embedding}. Inspired by~\cite{wang2017multi}, we adopt a RankNet loss function to learn the ranking relationships for instance $\mathcal{I}_{i}$:
\begin{equation}\label{ranknet}
\begin{split}
  \mathcal  L_{rank} = w_{i} \cdot & \Big( \mathop{ \sum_{p \in \mathcal{C}_{i}^{{+}}}^{} \sum_{q \in \mathcal{C}_{i}^{{-}}}^{} \log(1+\exp(o_{iq}-o_{ip}))} + \\
                &  \mathop{\sum_{j \in \mathcal{C}^{}}^{}\log(1+\exp(-\psi_{ij} o_{ij}))} \Big ),
\end{split}
\end{equation}
where $\mathcal{C}_{i}^{{+}}$ and $\mathcal{C}_{i}^{{-}}$ denote two sets of relevant and irrelevant labels to $i$-th instance. $\psi_{ij}$ is defined as an indicator function, where $\psi_{ij} = 1$ indicates that $i$-th instance is related to $j$-th label and
$\psi_{ij} = -1$ indicates that $i$-th instance is irrelative to $j$-th label.  $w_{i} = ( | \mathcal{C}_{i}^{{+}} | \cdot | \mathcal{C}_{i}^{{-}}| + | \mathcal{C}| )^{-1} $ plays a regularization role.
In the bracket of Eq.~(\ref{ranknet}), the first term gives punishment to the situation when the labels irrelevant to $\mathcal{I}_{i}$ have higher ranking orders than the relevant ones. The second term is used to enlarge the relatedness scores of the relevant pairs and reduce those of the irrelevant pairs.

Based on the above-trained model, the pairwise relatedness scores for the visual embeddings of target images and the semantic embeddings of target classes can be calculated. We rank the scores $\{o_{i1}, o_{i2}, \cdots , o_{iL_{t}}\}$ ($L_{t}$ is the number of unseen labels) in descending order, and select the classes of top-$k$ highest scores as the predicted target labels.

\subsection{Hash Code Learning}
For efficient nearest neighbor search, the semantic similarity of original images should be preserved in the Hamming space.
Generally, the similarity relationships can be defined with image labels. For a multi-label dataset, if two images share at least one label, they are considered similar, and dissimilar otherwise.
Let $\mathcal{B}$ be a set of hash codes for all images, and $\mathcal{S}$=$\{s_{ij} \} $ be the pairwise similarity matrix, then the conditional
probability of $s_{ij}$ can be defined as,
\begin{equation}
\label{eq:condition_prob}
p(s_{ij}|\mathcal{B})=
\begin{cases}
 \sigma (\Omega _{ij}) , & \ s_{ij} = 1, \\
 1 - \sigma (\Omega _{ij}) , & \ s_{ij} = 0,  \\
\end{cases}
\end{equation}
where $\Omega _{ij} =  \langle b_{i},b_{j} \rangle$ is the inner product of hash codes $b_{i}$ and $b_{j}$, and $\sigma (x) = \frac{1}{1+e^{-x}}$ is the sigmoid function, which scales the inner product value within [0, 1].

We adopt negative log-likelihood as the cost function to measure the pairwise similarity loss, as formulated by Eq.~(\ref{eq:l1}),
\begin{equation}
\label{eq:l1}
\begin{split}
\mathcal L_{p} & = - \sum_{s_{ij} \in \mathcal{S}}^{} log(p(s_{ij}|\mathcal{B})) \\
&= - \sum_{s_{ij} \in \mathcal{S}}^{} ( s_{ij} \cdot log(\sigma(\Omega _{ij}))+(1 - s_{ij}) \cdot log(1-\sigma (\Omega _{ij})) ).
\end{split}
\end{equation}
As $\sigma(\Omega _{ij})$=$\frac{1}{1+e^{-\Omega _{ij}}}$, the Eq.~(\ref{eq:l1}) can be rewritten as
\begin{equation}
\label{eq:l2}
\mathcal L_{p} = \sum_{s_{ij} \in \mathcal{S}}^{} ( log(1+ e^{\Omega _{ij}})- s_{ij} \cdot \Omega _{ij} ).
\end{equation}

It is very challenging to directly optimize this discrete optimization problem, as the binary constraint $b_{i} \in {\{-1,1\}^{M}}$ requires thresholding on the network outputs which may result in the vanishing-gradient problem in backpropagation.
We adopt the continuous relaxation strategy~\cite{liu2016deep,zhu2016deep} to solve this problem.
The output of deep hashing layer $u_{i}$ is fed to a tanh function $h_{i}= \tanh(u_{i})$, which is used as a substitute for binary code $b_{i}$.
Thus, $\Omega _{ij}$ is redefined as $h_{i}^{T}h_{j}$.

For more efficient and faster hash learning, we design a focal quantization loss to mitigate the divergence between the discrete binary codes and the continuous output of hashing networks, inspired by~\cite{lin2018focal}.
Since the gradient accumulations of a large number of simple samples are not helpful for training, the focal loss attempts to reduce the weights of simple examples to promote the training process.
First, we convert the binary code quantization problem into a binary classification problem.
We use a sigmoid activation to map the outputs of the hash layer into a probability distribution $\hat{p}_{i} = \sigma(u_{i})$.
Notice that, tanh and sigmoid are both monotonic increasing functions that hold the same variation trend, \textit{i.e.}, when $h_{i}$ asymptotically approaches to -1, $p_{i}$ also approaches 0, and vice versa (both approach to 1).
Thus the probability of binary classification can reflect the compactness of hash codes effectively.

The focal quantization loss is defined as
\begin{equation}
\label{eq:Q}
\begin{split}
\mathcal L_{q} = - \frac{1}{N} \sum_{i \in N}^{} \sum_{j \in M}^{}  ( & \hat{y}_{ij} \cdot (1-\hat{p}_{ij})^{\alpha} \cdot \log(\hat{p}_{ij}) + \\
        &  (1-\hat{y}_{ij}) \cdot (\hat{p}_{ij})^{\alpha} \cdot \log(1-\hat{p}_{ij})),
\end{split}
\end{equation}
where $\hat{y}_{i}$ is a label indicator that indicates which class (0 or 1) the output of hash layer should be classified as. We adopt a weighted sigmoid function to achieve such effect, \textit{i.e.},
$\hat{y}_{i} = \sigma(\beta \cdot u_{i})$, $\beta$ is a parameter far greater than 1.

By integrating the pairwise similarity loss and focal quantization loss, the overall hashing loss can be defined as
\begin{equation}
\label{eq:C}
\mathcal L_{hash} =  \mathcal L_{p} + \mathcal L_{q}.
\end{equation}

\textit{}\section{Experiments}
\label{sec:experiment}

\subsection{Datasets}
To verify the performance of the proposed method, we compare the proposed method with several baselines on three widely used multi-label image datasets, \textit{i.e.}, {NUS-WIDE}, {VOC2012}, and {COCO}.

\textbf{NUS-WIDE}~\cite{chua2009nus} is a dataset containing 269,648 public web images. Each image is annotated with one or more class labels from a total of 81 classes. There exists a widely used subset of images associated with the 21 most common labels and each label associated with at least 5,000 images, resulting in a total of 195,834 images.

\textbf{VOC2012}~\cite{everingham2010pascal} is a widely used dataset for object detection and segmentation, which contains 17,125 images. Each image is associated with at least one of the 20 semantic labels.

\textbf{COCO}~\cite{Tsung2014coco} is a dataset for object detection, semantic scene labeling, and indexing, which contains 123,287 images with semantic labels. Each image is associated with one to sixteen labels from a total of 90.

\subsection{Implementation Details}

To construct a zero-shot scenario, we should further split the dataset{\footnote{https://github.com/qinnzou/Zero-Shot-Hashing}. Since there are more complex semantic relationships among these multi-label datasets, we use one of these three image datasets as source data and one as target data. For example, we can set NUS-WIDE as source data and VOC2012 as target data, and vise versa. Before training models based on these datasets, data preprocessing must be done. We  set three experiments, including one between NUS-WIDE and VOC2012, one between NUS-WIDE and COCO, and the last one between COCO and VOC2012.

\subsubsection{Experiment between NUS-WIDE and VOC2012}~ In NUS-WIDE, we remove the common concepts (semantic labels) shared by these two datasets and related images, because there are much more images in NUS-WIDE than in VOC2012. In VOC2012, we remove several ambiguous concepts and related images. Such data-clean operations result in a subset of NUS-WIDE containing 106,389 images and 18 labels, and a subset of VOC2012 containing 16,750 images and 17 labels. For NUS-WIDE, we randomly select 10,000 images as the training set, 2,000 images as the test query set, and the rest as the retrieval database. For VOC2012, we randomly select 4,000 images as the training set, 1,000 images as the test query set, and the rest as the retrieval database.

\subsubsection{Experiment between NUS-WIDE and COCO}~ We remove the common concepts and relative images from NUS-WIDE and keep COCO unchanged. Finally, a subset of NUS-WIDE containing 100,303 images and 17 labels, and a subset of COCO containing 123,274 images and 80 labels are prepared for the following experiments. For both datasets, we randomly select 10,000 images as the training set, 2,000 images as the test query set, and the rest as the retrieval database.

\subsubsection{Experiment between VOC2012 and COCO}~ We remove the images related to three ambiguous concepts for VOC2012. After that, all the concepts of VOC2012 are included in COCO. Then, for COCO, we remove the images that have the concepts in VOC2012. Finally, there remains 21,987 images and 60 labels for COCO, and 16,750 images and 17 labels for VOC2012. For both datasets, we randomly select 4,000 images for training, 1,000 images as the test query, and the rest as the retrieval database.

For NUS-WIDE, VOC2012, and COCO, we check the labels and ensure that the training/query set contains all the concepts of the corresponding dataset.

\begin{figure}[!t]
	\centering
	\subfigure[\scriptsize NUS-WIDE $\rightarrow$ VOC2012]{
		\includegraphics[width=0.45\linewidth]{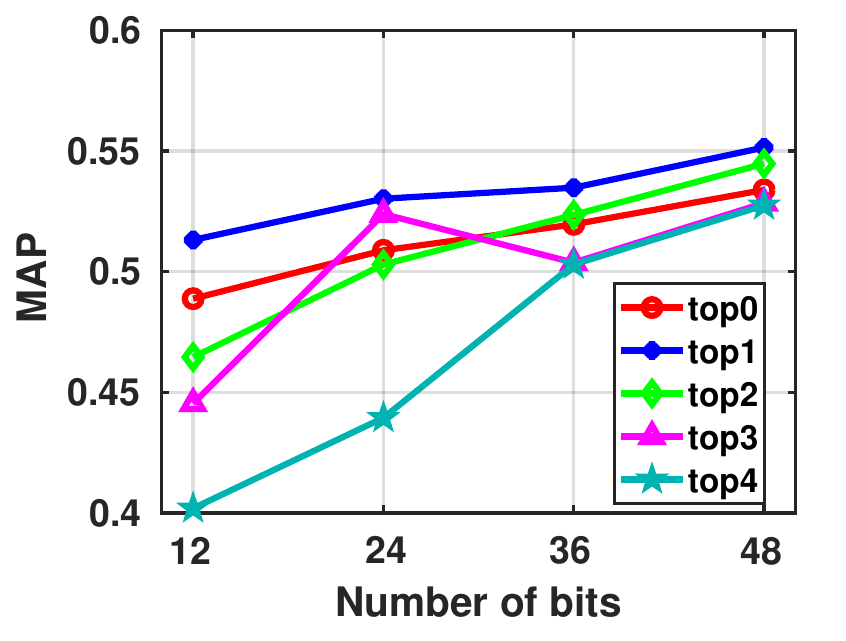}
	}
	\subfigure[\scriptsize VOC2012 $\rightarrow$ NUS-WIDE]{
		\includegraphics[width=0.45\linewidth]{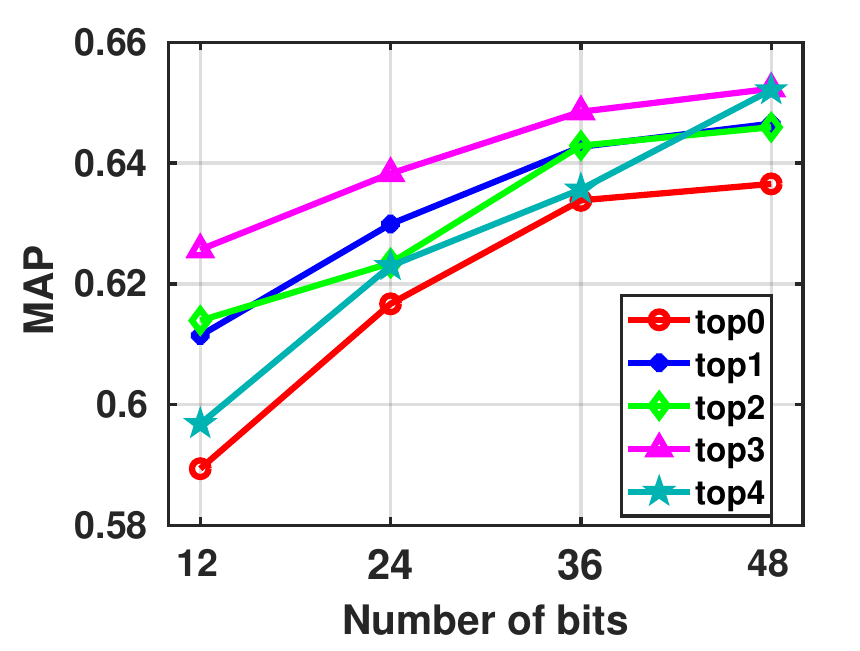}
	}
	\subfigure[\scriptsize NUS-WIDE $\rightarrow$ COCO]{
		\includegraphics[width=0.45\linewidth]{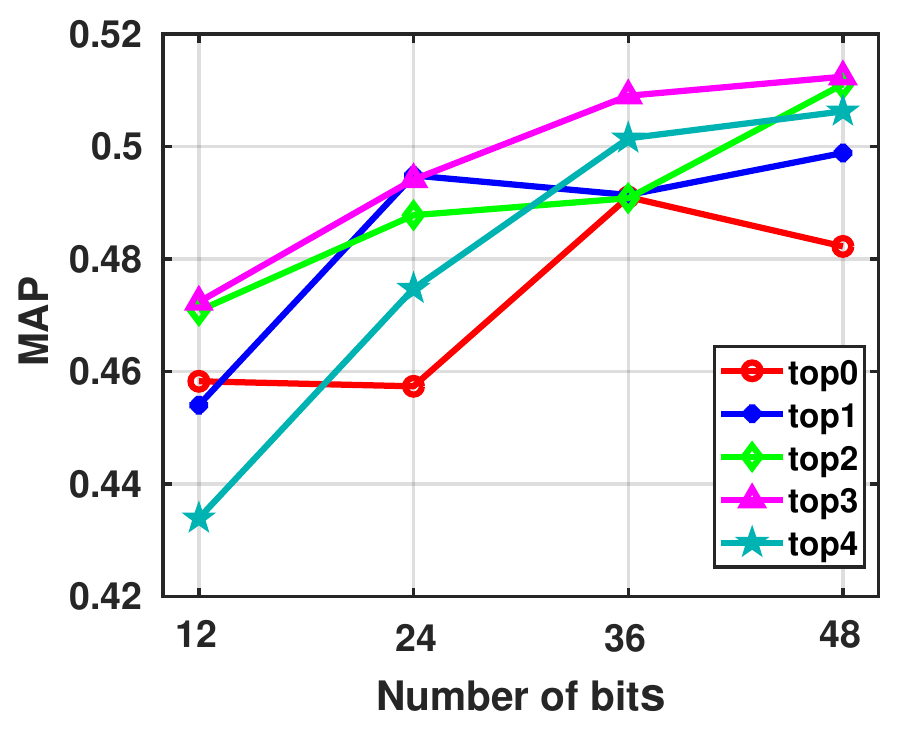}
	}
	\subfigure[\scriptsize COCO $\rightarrow$ NUS-WIDE]{
		\includegraphics[width=0.45\linewidth]{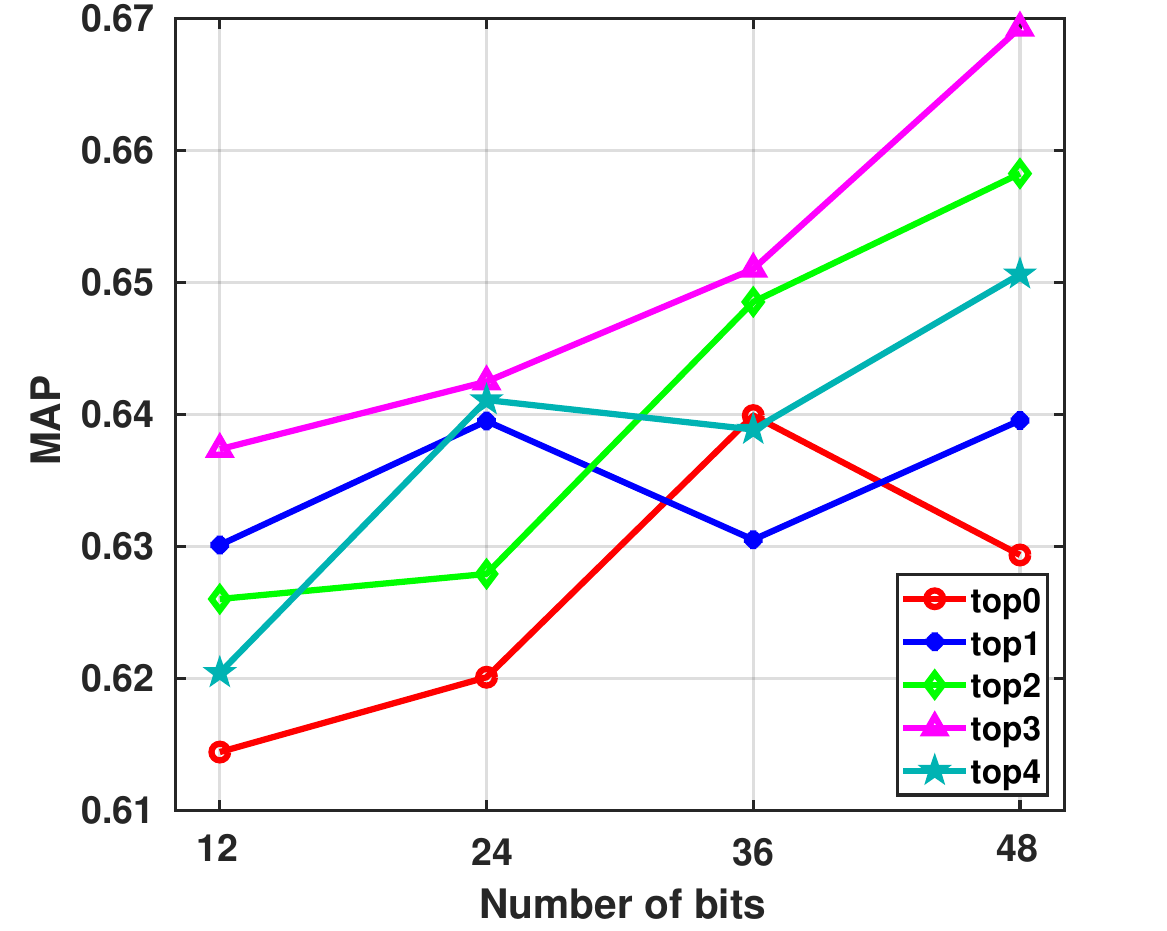}
	}
	\subfigure[\scriptsize{VOC $\rightarrow$ COCO}]{
		\includegraphics[width=0.45\linewidth]{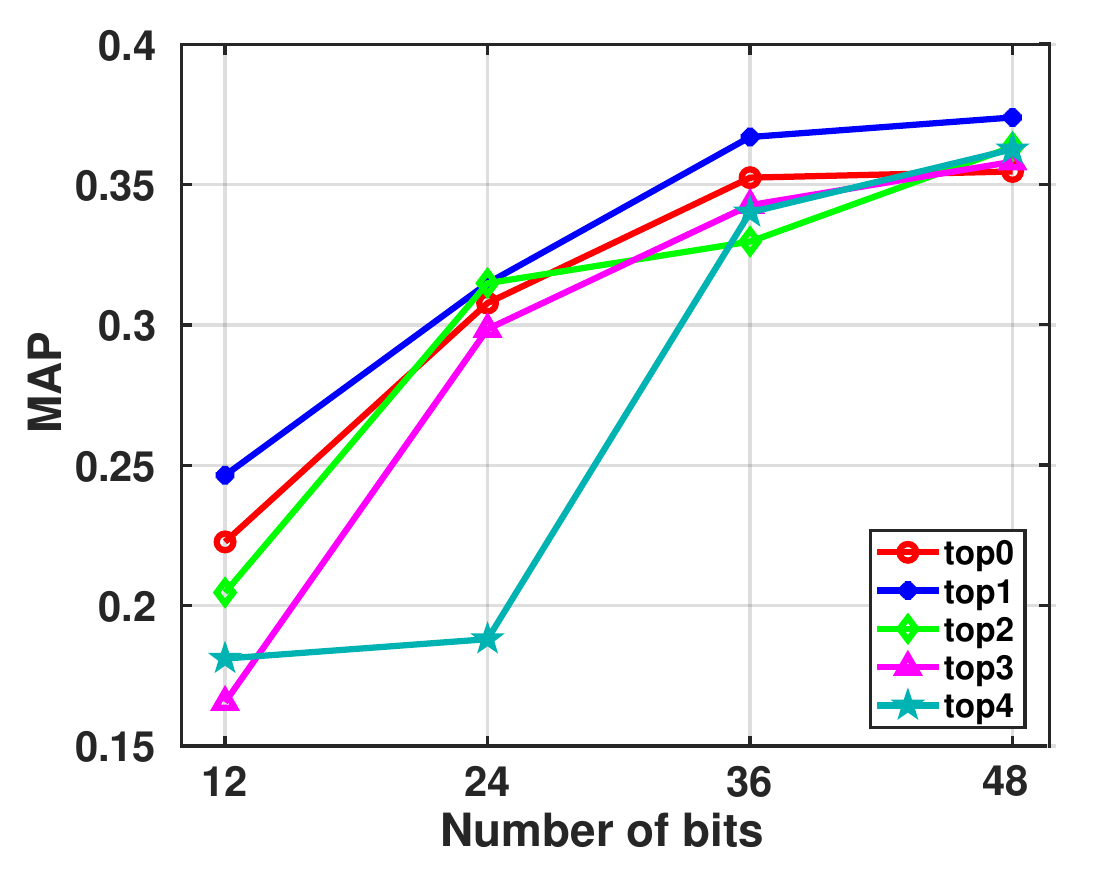}
	}
	\subfigure[\scriptsize {COCO $\rightarrow$ VOC}]{
		\includegraphics[width=0.45\linewidth]{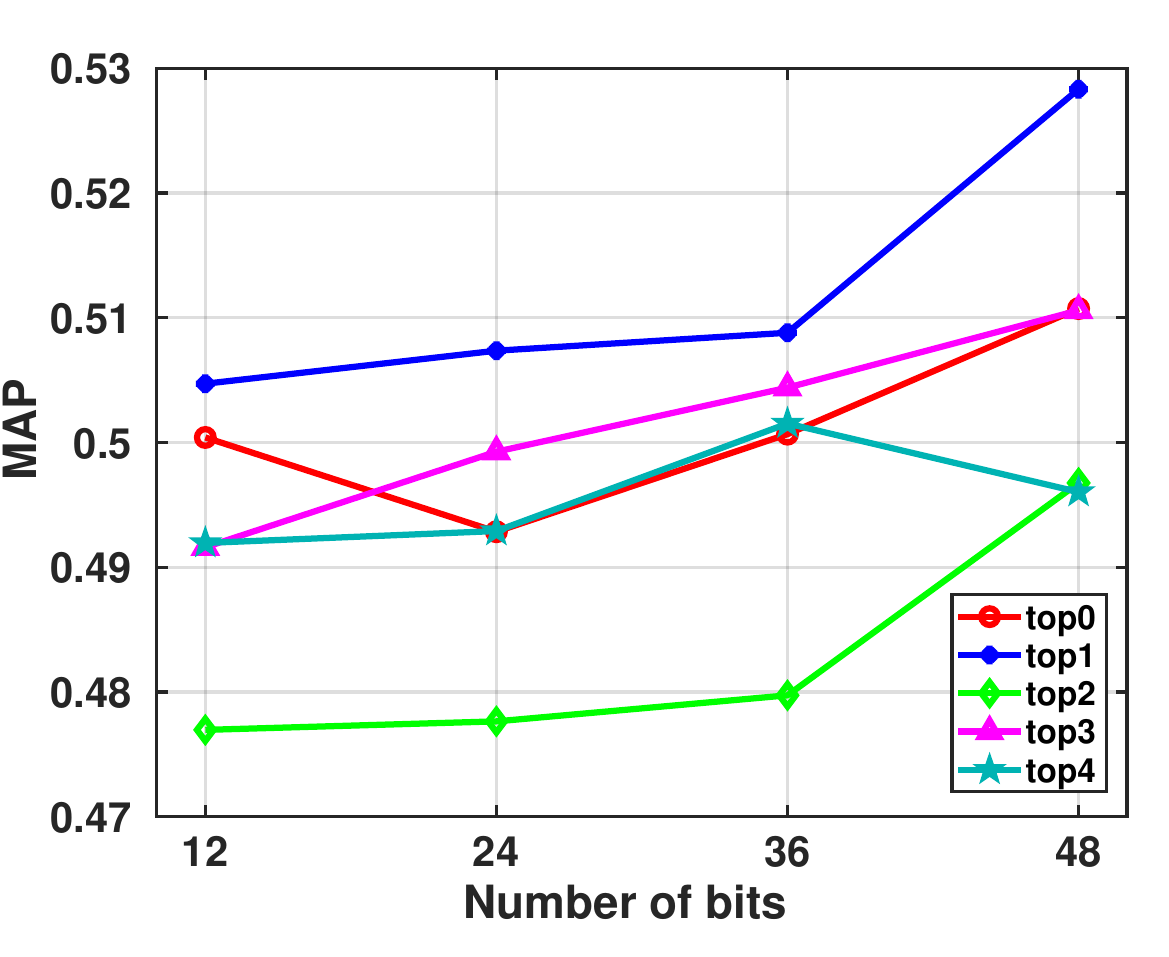}
	}
	\caption{Performance obtained by using different numbers of predicted labels for target data with hash codes of 12, 24, 36, and 48 bits, respectively. The values are computed based on the top-1000 retrieved images.}
	\label{fig:setting}
\end{figure}

We implement the proposed method (T-MLZSH) using the TensorFlow toolkit. In this paper, we use AlexNet as the backbone CNN. To validate the versatility of the proposed framework, we will also evaluate by replacing the backbone CNN with VGG16 and ResNet50. We use the pre-trained model to initialize the weight parameters and focus on training the hashing layer and embedding layer. Adam method is adopted for stochastic optimization with a mini-batch size of 128, and all input images are resized to 227$\times$227.

We compare our method (T-MLZSH) with nine other methods, including the traditional methods KSH~\cite{liu2012supervised}, IMH~\cite{shen2013inductive}, SDH~\cite{shen2015supervised}, ZSH-DA~\cite{pachori2018hashing}, and ZSH~\cite{yang2016zero}, and the deep learning-based methods DHN~\cite{zhu2016deep}, Hashnet~\cite{cao2017hashnet}, ADSH~\cite{adsh}, and TZSH~\cite{lai2018transductive}.
Among the traditional methods, ZSH-DA and ZSH are two zero-shot hashing methods. Among the deep learning-based methods, TZSH is a transductive zero-shot hashing method. It is worth noting that, all the methods use the same training and test settings on the newly formed datasets.

In the training, for KSH, IMH, SDH, DHN, Hashnet, and ADSH, the images and labels of the training set are used, while for ZSH-DA, ZSH, TZSH, and the proposed T-MLZSH, in addition to the images and labels in the training set, the raw images (without labels) in the test query set are also used.
Note that, ADSH is a supervised method which treats the query points and database points in an asymmetric way, and DIHN~\cite{DBLP:conf/cvpr/WuDL0W19} successfully extends the asymmetric training strategy in an incremental learning framework. As DIHN uses the labels of the unseen test query data for training, which violates the protocols of our experiments, we do not include it in the comparison.

For deep learning-based methods, we use the raw images as input. For the non-CNN approaches, we use the outputs of the fc7 layer in AlexNet as their visual features. All the experiments are conducted on an NVIDIA TITAN Xp GPU and the Ubuntu 16.04 operating system.

\subsection{Metrics}
The metrics we used to evaluate the image retrieval quality are four widely-used metrics: Average Cumulative Gains (ACG), Normalized Discounted Cumulative Gains (NDCG), Mean Average Precision(MAP), and Weighted Mean Average Precision(WAP)~\cite{cao2011cvpr,jarvelin2002cumulated,zhao2015deep,zou2019hash}.

MAP is the mean of average precision for each query, which
can be calculated by
\begin{equation}\label{eq:MAP}
	MAP=\frac{1}{Q}\sum_{Q}^{q}AP\left ( q \right )
\end{equation}

where
\begin{equation}\label{AP}
	AP(q)=\frac{1}{N_{tr}(q)@n}\sum_{n}^{}{i}(Tr(q,i)\frac{N_{tr}(q)@i}{i})
\end{equation}
$Tr(q,i)\in {0,1}$is an indicator function that if $I_{q}$ and $I_{i}$ have same class labels, $Tr(q,i)=1$;otherwise $Tr(q,i)=0$. Q is the number of query sets and $N_{tr}(q)@i$ indicates the number of relevant images w.r.t the query image $I_{q}$ within the top $i$ images.
ACG represents the average number of shared labels between the query image and the top $n$ retrieved images. For a given query image $I_{q}$, the ACG score of the top $n$ retrieved images is calculated by
\begin{equation}\label{ACG}
	ACG@n=\frac{1}{n}\sum_{n}^{i}C(q,i)
\end{equation}
where n denotes the number of top retrieval images and $C(q,i)$ is the number of shared class labels between $I_{q}$ and $I_{i}$.
NDCG is a popular evaluation metric in information retrieval. Given a query image $I_{q}$, the DCG score of top $n$ retrieved images is defined as
\begin{equation}\label{DCG}
	DCG@n=\sum_{n}^{i}\frac{2^{C(q,i)}-1}{log(1+i)}
\end{equation}
Then, the normalized DCG(NDCG) score at the position n can be calculated by $NDCG@n==\frac{DCG@n}{Z_{n}}$, where $Z_{n}$ is the maximum value of DCG@n, which constrains the value of NDCG in the range [0,1].

WAP is similar to MAP, the only difference is that WAP is the average value of ACG scores at each tip $n$ retrieval image rather than average precision. WAP can be calculated by
\begin{equation}\label{eq:WAP}
	WAP=\frac{1}{Q}\sum_{Q}^{q}(\frac{1}{N_{Tr}(q)@n}\sum_{i}^{n}(Tr(q,i)\times ACG@i))
\end{equation}

\subsection{Overall Performance}
In this part, we analyze the retrieval results all evaluated on the unseen target data.

\renewcommand\arraystretch{1.2}
\begin{table*}[!t]
	\centering
    \caption{MAP results obtained by using different numbers of bits on (NUS-WIDE, VOC2012). The MAPs are computed based on the top-1000 retrieved images.}
	\small
	{
		\begin{tabular}{|c|cccc|cccc|}
			\hlinew{1.2pt}
			\multirow{2}{*}{Methods} &
			\multicolumn{4}{c|}{NUS-WIDE $\rightarrow$ VOC2012} &
			\multicolumn{4}{c|}{VOC2012 $\rightarrow$ NUS-WIDE}
			\\
			\cline{2-5}
			\cline{6-9}
			& 12-bit & 24-bit & 36-bit & 48-bit & 12-bit & 24-bit & 36-bit & 48-bit \\
			\hlinew{1.0pt}
			\multirow{1}{*}{KSH~\cite{liu2012supervised}} & 0.4033 & 0.4079 & 0.4153 & 0.4181 & 0.5476 & 0.5510 & 0.5602 & 0.5670   \\
			\multirow{1}{*}{SDH~\cite{shen2015supervised}} & 0.4097 & 0.4010 & 0.4087 & 0.4095 & 0.5327 & 0.5345 & 0.5368 & 0.5395   \\
			\multirow{1}{*}{IMH~\cite{shen2013inductive}} & 0.4083 & 0.4346 & 0.4320 & 0.4297 & 0.5616 & 0.5723 & 0.5718 & 0.5710   \\
			\hline
			\multirow{1}{*}{DHN~\cite{zhu2016deep}} & 0.4171 & 0.4282 & 0.4362 & 0.4395 & 0.5664 & 0.5739 & 0.5726 & 0.5688    \\
			\multirow{1}{*}{Hashnet~\cite{cao2017hashnet}} & {0.4048} & {0.4288} & {0.4349} & 0.4465 & {0.5674} & {0.5940} & {0.6073} & {0.6336} \\
			\multirow{1}{*}{ADSH~\cite{adsh}} & {0.3988} & {0.4163} & {0.4060} &  {0.4201} &{0.5315} & {0.5473} & {0.5392} & {0.5853}   \\
			\hline
			\multirow{1}{*}{ZSH-DA~\cite{pachori2018hashing}} & 0.3592 & 0.3618 & 0.3770 & 0.3596 & 0.5132 & 0.5166 & 0.5212 & 0.5191   \\
			\multirow{1}{*}{ZSH~\cite{yang2016zero}} & 0.3968 & 0.4055 & 0.4111 & 0.4296  & 0.5340 & 0.5566 & 0.5589 & 0.5500   \\
			\multirow{1}{*}{TZSH~\cite{lai2018transductive}} & 0.4413 & 0.4683 & 0.4644 &  0.4753 & 0.5736 & 0.5805 & 0.5919 & 0.5896   \\
			\rowcolor{mygray1} \multirow{1}{*}{T-MLZSH} & \textbf{0.4808} & \textbf{0.4884} & \textbf{0.4894} & \textbf{0.5037} & \textbf{0.6106} & \textbf{0.6131} & \textbf{0.6149} & \textbf{0.6200}  \\
			\hlinew{1.2pt}
	\end{tabular}}
	\label{table:one}
\end{table*}

\begin{table*}[!t]
	\centering
\caption{MAP results obtained by using different numbers of bits on (NUS-WIDE, COCO). The MAPs are computed based on the top-1000 retrieved images.}
	\small
	{
		\begin{tabular}{|c|cccc|cccc|}
			\hlinew{1.2pt}
			\multirow{2}{*}{Methods} &
			\multicolumn{4}{c|}{NUS-WIDE $\rightarrow$ COCO} &
			\multicolumn{4}{c|}{COCO $\rightarrow$ NUS-WIDE}
			\\
			\cline{2-5}
			\cline{6-9}
			& 12-bit & 24-bit & 36-bit & 48-bit & 12-bit & 24-bit & 36-bit & 48-bit \\
			\hlinew{1.0pt}
			\multirow{1}{*}{KSH~\cite{liu2012supervised}} & 0.3948 & 0.4069 & 0.4113 & 0.4143 & 0.5948 & 0.6167 & 0.6191 & 0.6224   \\
			\multirow{1}{*}{SDH~\cite{shen2015supervised}} & 0.3782 & 0.3917 & 0.3971 & 0.4050 & 0.5681 & 0.5954 & 0.6102 & 0.6051   \\
			\multirow{1}{*}{IMH~\cite{shen2013inductive}} & 0.3905 & 0.4021 & 0.4114 & 0.4188 & 0.5983 & 0.5961 & 0.6098 & 0.6132   \\
			\hline
			\multirow{1}{*}{DHN~\cite{zhu2016deep}} & 0.4250 & 0.4325 & 0.4529 & 0.4487 & 0.6177 & 0.6421 & 0.6466 & 0.6559    \\
			\multirow{1}{*}{Hashnet~\cite{cao2017hashnet}} & 0.3908 & 0.4018 & 0.4129 &  {0.4439} & {0.5538} & {0.5674} & {0.5830} & {0.5932}   \\
			\multirow{1}{*}{ADSH~\cite{adsh}} & {0.3783} & {0.3890} & {0.3968} &  {0.4056} & {0.5773} & {0.5906} & {0.5871} & {0.6219}   \\
			\hline
			\multirow{1}{*}{ZSH-DA~\cite{pachori2018hashing}} & 0.3597 & 0.3592 & 0.3772 & 0.3744 & 0.5256 & 0.5220 & 0.5230 & 0.5247   \\
			\multirow{1}{*}{ZSH~\cite{yang2016zero}} & 0.3832 & 0.4091 & 0.4109 & 0.4286  & 0.5708 & 0.5727 & 0.5753 & 0.5782   \\
			\multirow{1}{*}{TZSH~\cite{lai2018transductive}} & 0.4436 & 0.4585 & 0.4660 &  0.4800 & 0.5933 & 0.6368 & 0.6070 & 0.6336   \\
			
			\rowcolor{mygray1} \multirow{1}{*}{T-MLZSH} & \textbf{0.4724} & \textbf{0.4941} & \textbf{0.5090} & \textbf{0.5124} & \textbf{0.6374} & \textbf{0.6425} & \textbf{0.6510} & \textbf{0.6693}  \\
			\hlinew{1.2pt}
	\end{tabular}}
	
	\label{table:two}
\end{table*}

\subsubsection{The number of predicted unseen categories}
Figure~\ref{fig:setting} displays the results of using different numbers of predicted labels on the target data. Top-$k$ indicates that the first $k$ categories in the correlation-score ranking list are used as predicted labels and top-0 means that the labels of target data are set to a vector of all zeros. From Fig.~\ref{fig:setting}(a) and (b), we can see that when setting VOC2012 as target data, using top-1 predicted labels can achieve the best performance. The possible reason is that the average number of objects in each image on VOC2012 is relatively small. With more predicted labels used for supervised hashing learning, it will inevitably incur misleading information and cause performance degradation. When setting NUS-WIDE as target data, the best results are obtained by using top-3 predicted labels. In the following experiments, we use top-1 and top-3 predicted labels as supervised information for the proposed method in default for VOC2012 and NUS-WIDE, respectively.
In Fig.~\ref{fig:setting}(c)-(f), we can see that the best performance can be achieved by using top-3 predicted labels for the experiments on COCO and NUS-WIDE, and using top-1 for experiments on COCO and NUS-WIDE. It may be because that the average number of objects in each image in the (NUS-WIDE, COCO) case is 2.48 and 2.97,  and that in the (VOC, COCO) case is 1.30 and 1.48, respectively. Hence, we use top-3 predicted labels as supervised information for the proposed method in default for (COCO, NUS-WIDE) and top-1 for (VOC, COCO) in the following experiments.

\subsubsection{Results under different scenarios}
The MAP results of the proposed method and the comparison methods are shown in Table~\ref{table:one},~\ref{table:two}, and~\ref{table:three}. It can be seen that the non-zero-shot deep hashing methods DHN, Hashnet, and ASDH outperform the traditional hashing methods KSH, IMH, SDH, ZSH-DA, and ZSH. Among the non-zero-shot deep hashing methods, ADSH has lower performance than the other two. The possible reason is that it treats the query points and database points in an asymmetric way which drives the network to pay more attention to the seen data. Thus, when the dataset is replaced by the unseen dataset, ADSH's performance will be affected. The two traditional zero-shot hashing methods ZSH and ZSH-DA achieve low performance on the multi-label datasets, which indicates that the complex semantics of multi-label images are too hard to be modeled by learning a one-to-one semantic representation. The zero-shot deep hashing methods T-MLZSH and TZSH obtain significantly higher performance than the other methods.

It can also be seen that the proposed T-MLZSH outperforms almost all the comparison methods significantly on all target datasets under different scenarios. On NUS-WIDE and VOC2012, the transductive zero-shot hashing methods, \textit{i.e.}, TZSH and T-MLZSH, achieve higher MAP values than other methods, as shown in Table~\ref{table:one}. Compared to TZSH, T-MLZSH achieves increments of about 3.1\% and 2.8\% in the average MAP for different bits on NUS-WIDE and VOC2012, respectively. The possible reason is that TZSH adopts a strategy only utilizing partially-selected target data for hash learning, which limits its performance.

From Table~\ref{table:two} we can see, compared to TZSH, T-MLZSH achieves increments of about 3.4\% or 3.2\% in the average MAP for different bits on COCO$\rightarrow$NUS-WIDE and NUS-WIDE$\rightarrow$COCO, respectively. In this experiment, the deep supervised hashing method DHN is found to perform better than TZSH on COCO$\rightarrow$NUS-WIDE. The possible reason is that COCO is categorized into more categories and DHN can get more detailed supervised information when setting COCO as the training set. Nevertheless, T-MLZSH still outperforms DHN by about 0.95\% in the average MAP.

\begin{table*}[!t]
    \caption{MAP results obtained by using different numbers of bits on (VOC2012, COCO). The MAPs are computed based on the top-1000 retrieved images.}
	\centering
	\small
	{
		\begin{tabular}{|c|cccc|cccc|}
			\hlinew{1.2pt}
			\multirow{2}{*}{Methods} &
			\multicolumn{4}{c|}{VOC2012 $\rightarrow$ COCO} &
			\multicolumn{4}{c|}{COCO $\rightarrow$ VOC2012}
			\\
			\cline{2-5}
			\cline{6-9}
			& 12-bit & 24-bit & 36-bit & 48-bit & 12-bit & 24-bit & 36-bit & 48-bit \\
			\hlinew{1.0pt}
			\multirow{1}{*}{KSH~\cite{liu2012supervised}} & 0.1763  & 0.2076  & 0.2346  & 0.2490 & 0.4498 & 0.4781 &0.4912 & 0.5005   \\
			\multirow{1}{*}{SDH~\cite{shen2015supervised}} &0.1444   & 0.1762  &0.1871   & 0.1991 &0.4656  &0.4869  &0.5045 & 0.5060 \\
			\multirow{1}{*}{IMH~\cite{shen2013inductive}} &0.2005   & 0.2151  & 0.2279  &0.2317  &0.4083  &0.4291  &0.4347 & 0.4380  \\
			\hline
			\multirow{1}{*}{DHN~\cite{zhu2016deep}} &0.1635  &0.1826   &0.1954   &0.2351  &0.4311  &0.4370  &0.4573 &0.4597    \\
			\multirow{1}{*}{Hashnet~\cite{cao2017hashnet}} & 0.1521 & 0.2029 & 0.2470 &  0.2585 & 0.4331 & 0.4605 & 0.4766 & 0.4924   \\
			\multirow{1}{*}{ADSH~\cite{adsh}} & 0.1368 & 0.1472 & 0.1771 &  0.1884 & 0.4310 & 0.4794 & 0.5018 & 0.5061   \\
			\hline
			\multirow{1}{*}{ZSH-DA~\cite{pachori2018hashing}} & 0.0853  &0.0746   & 0.0709  & 0.1104 &0.3924  & 0.4189 & 0.4247 & 0.4393 \\
			\multirow{1}{*}{ZSH~\cite{yang2016zero}} & 0.1602  &0.1935   &0.2249   &0.2227  &0.4658  &0.4775  &0.5003 &0.4981   \\
			\multirow{1}{*}{TZSH~\cite{lai2018transductive}} & 0.2336 &0.2632 &0.2690 &0.2631   &0.4967 	&0.5053 &0.5128	&  0.5079  \\
			
			\rowcolor{mygray1} \multirow{1}{*}{T-MLZSH} & \textbf{0.2465} & \textbf{0.3149} & \textbf{0.3670} & \textbf{0.3740} & \textbf{0.5047} & \textbf{0.5276} & \textbf{0.5296} & \textbf{0.5345}  \\
			\hlinew{1.2pt}
	\end{tabular}}
	\label{table:three}
\end{table*}

\begin{figure*}[!t]
	\centering
		\subfigure[\small NUS-WIDE $\rightarrow$ VOC2012]{
		\includegraphics[width=0.45\linewidth]{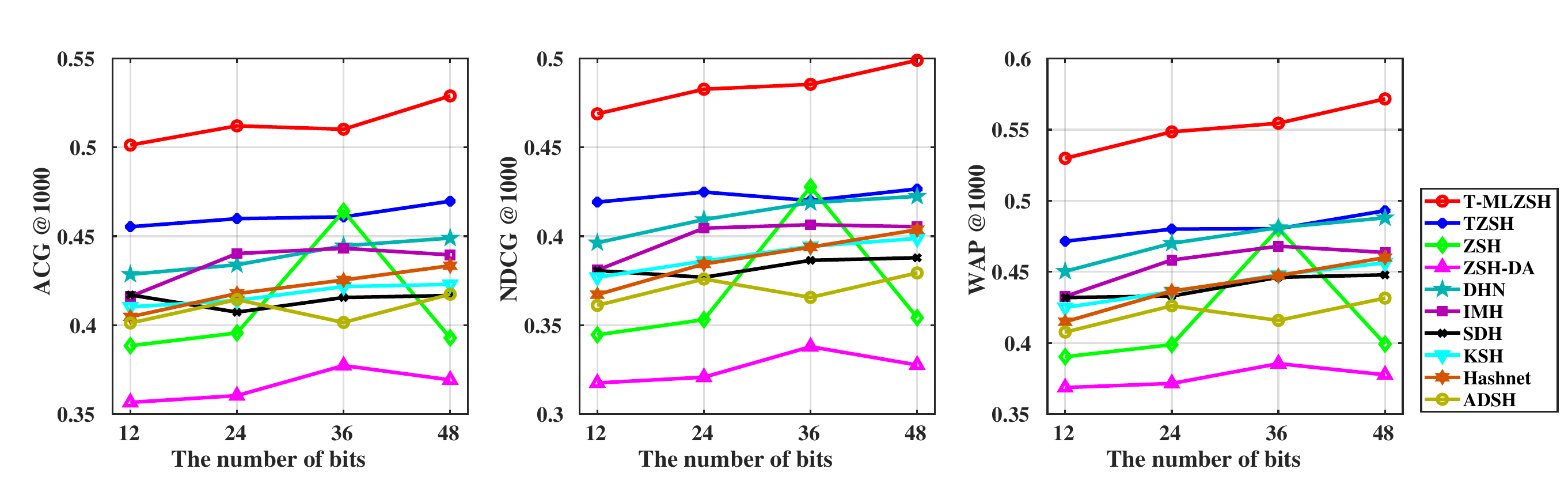}
	}
	\subfigure[\small VOC2012 $\rightarrow$ NUS-WIDE]{
		\includegraphics[width=0.45\linewidth]{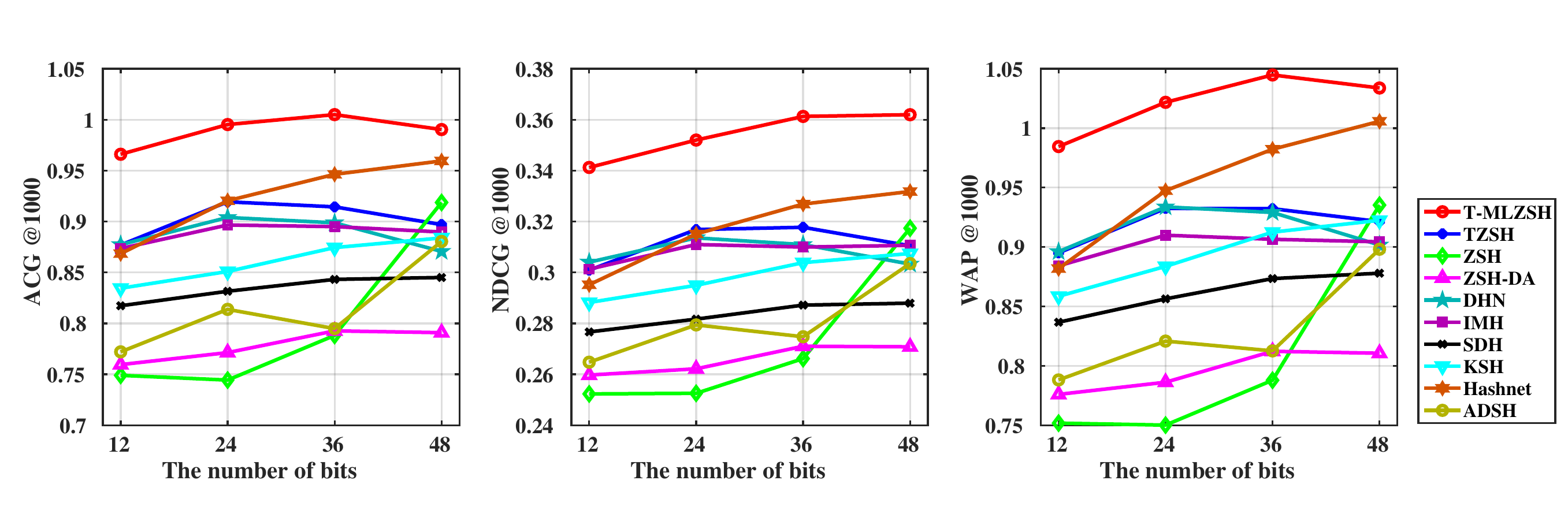}
	}
	\subfigure[\small NUS-WIDE $\rightarrow$ COCO]{
		\includegraphics[width=0.45\linewidth]{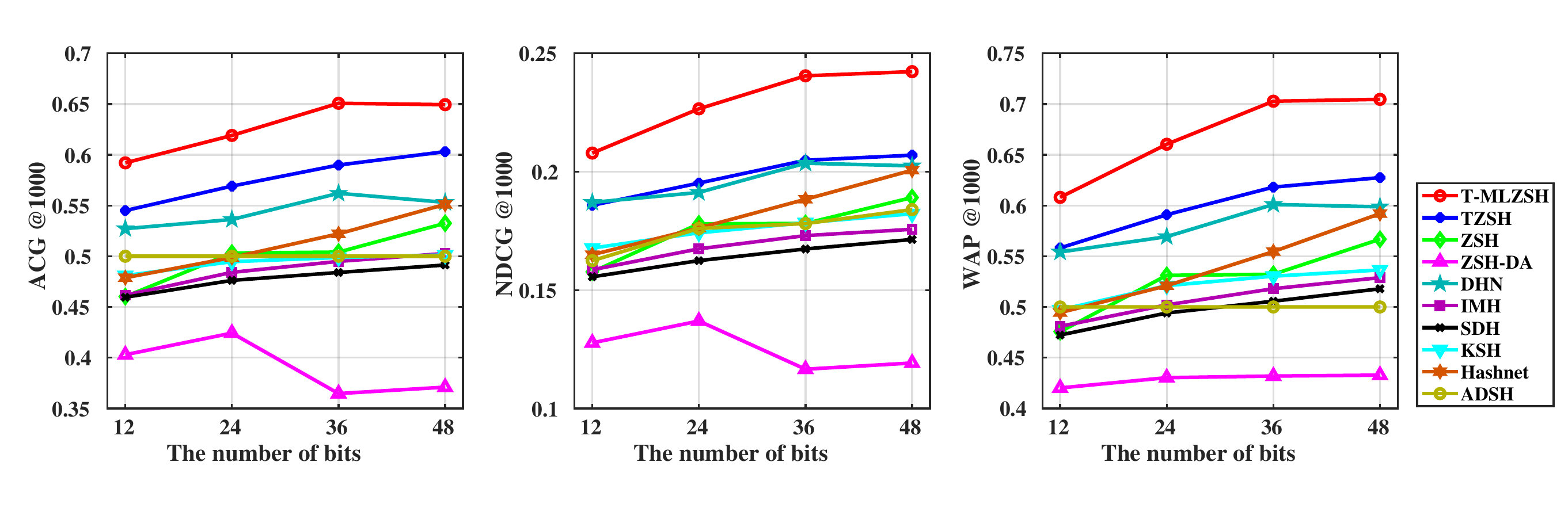}
	}
	\subfigure[\small COCO $\rightarrow$ NUS-WIDE]{
		\includegraphics[width=0.45\linewidth]{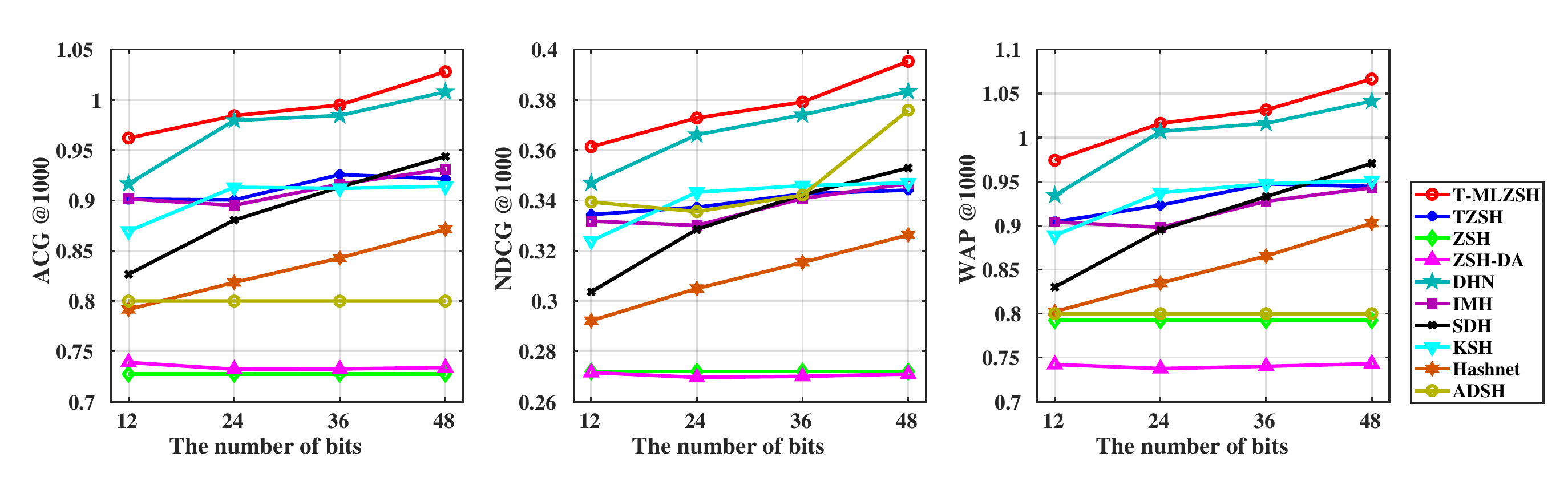}
	}

	\subfigure[\small {VOC2012 $\rightarrow$ COCO}]{
		\includegraphics[width=0.45\linewidth]{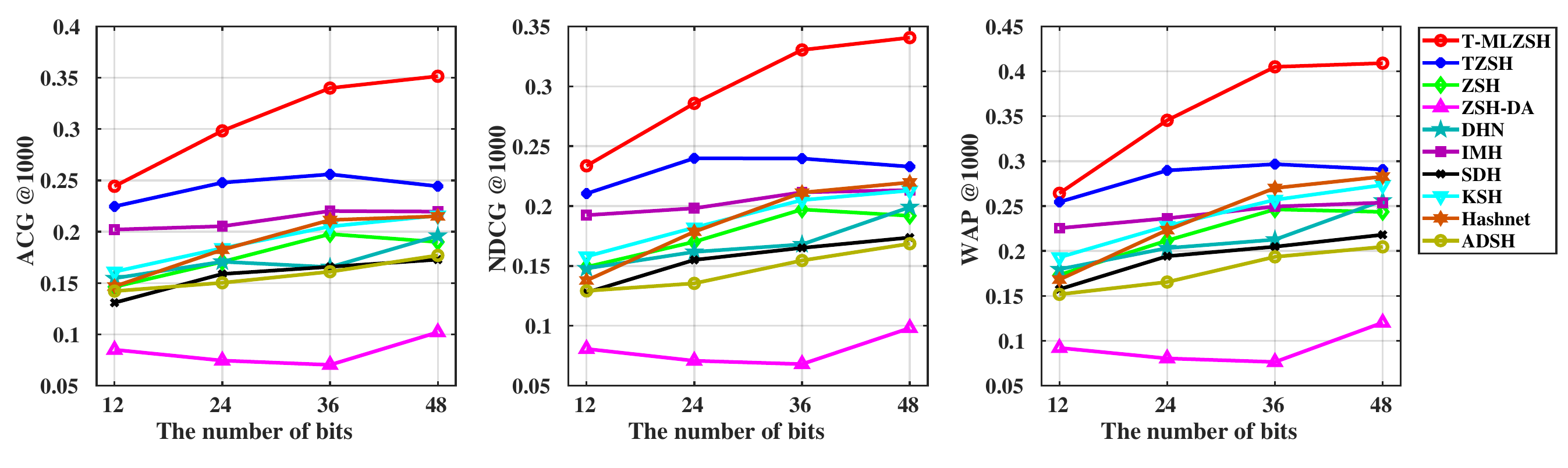}
	}
	\subfigure[\small {COCO $\rightarrow$ VOC2012}]{
		\includegraphics[width=0.45\linewidth]{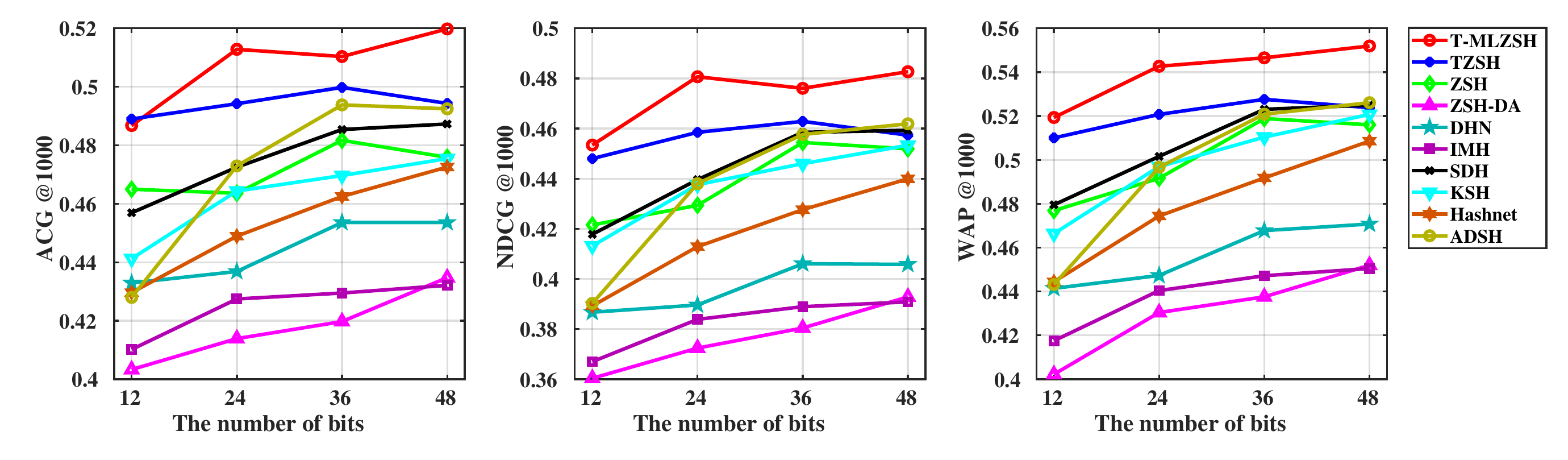}
	}
	\caption{Comparison of results obtained by the proposed method on ACG, NDCG, and WAP with 12-, 24-, 36-, and 48-bits hash codes, with regarding  the top-1000 retrieved images.}
	\label{fig:metric-coco-nus}
\end{figure*}

Table~\ref{table:three} shows the results on (COCO, VOC2012). Compared to TZSH, T-MLZSH achieves increments of about 6.8\% or 1.8\% on average MAP for different bits. The proposed method is found to obtain a significantly better performance on VOC2012$\rightarrow$COCO, where COCO is unseen. The possible reason is that, although all the instances of VOC2012 are included in COCO, some unlabeled targets in COCO images are similar to that in VOC2012. Features extracted from VOC2012 images may possess high similarity with some COCO images. Since the scale of COCO is much larger than VOC2012, the overall performance is lower when COCO is unseen.

More results in the other three metrics, \textit{i.e.}, ACG, NDCG, and WAP, are presented in  Fig.~\ref{fig:metric-coco-nus}. We can see that the overall trends of the performance on the three metrics are consistent.  When the code length increases, the performance improves. According to the definition, these three metrics can make a more fair evaluation on multi-label images, as the numbers of shared labels between images are considered. For MAP, the pairwise images that share at least one common object label will be considered as relevant images, and no more comparisons of fine-grained semantic relation between these images are included, which may not stay in step with user demand in multi-label image
retrieval. For WAP, the average number of shared class labels among these retrieved similar images is considered. Higher WAP means more high-quality retrieval results having more shared class labels in the nearest search. Although the range of WAP on different datasets would be different, the WAPs of T-MLZSH are stably higher than that of the comparison methods.

\begin{figure*}[!t]
	\centering
	\includegraphics[width=0.98\linewidth]{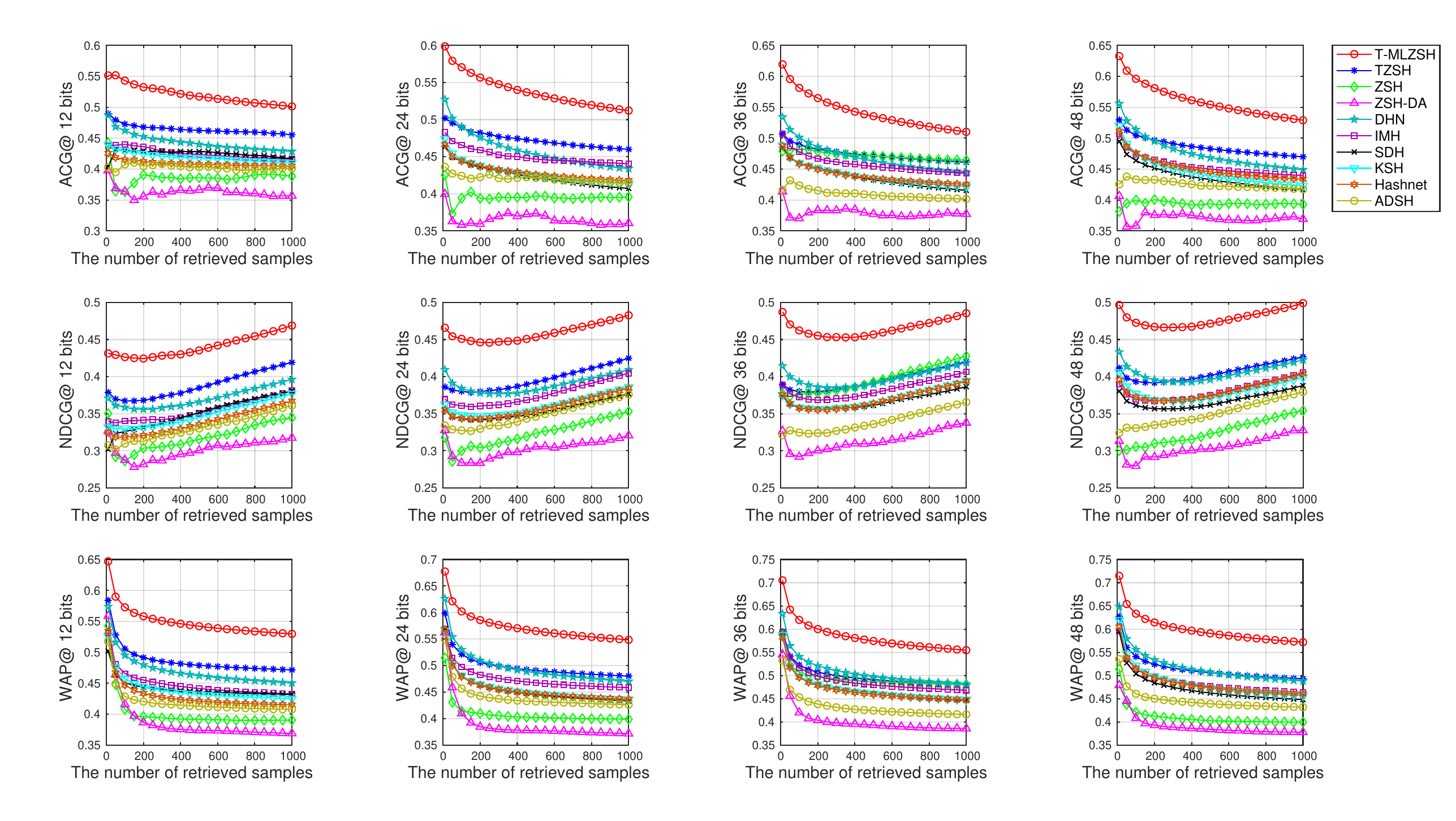}\\
	\caption{Performance comparison on NUS-WIDE$\rightarrow$VOC2012. The VOC2012 dataset is unseen. From top to bottom, there are ACG, NDCG, and WAP w.r.t. different top returned
		samples with hash codes of 12, 24, 36, and 48 bits, respectively.}
	\label{fig:pascal}
\end{figure*}

\begin{figure*}[!t]
	\centering
	\includegraphics[width=0.98\linewidth]{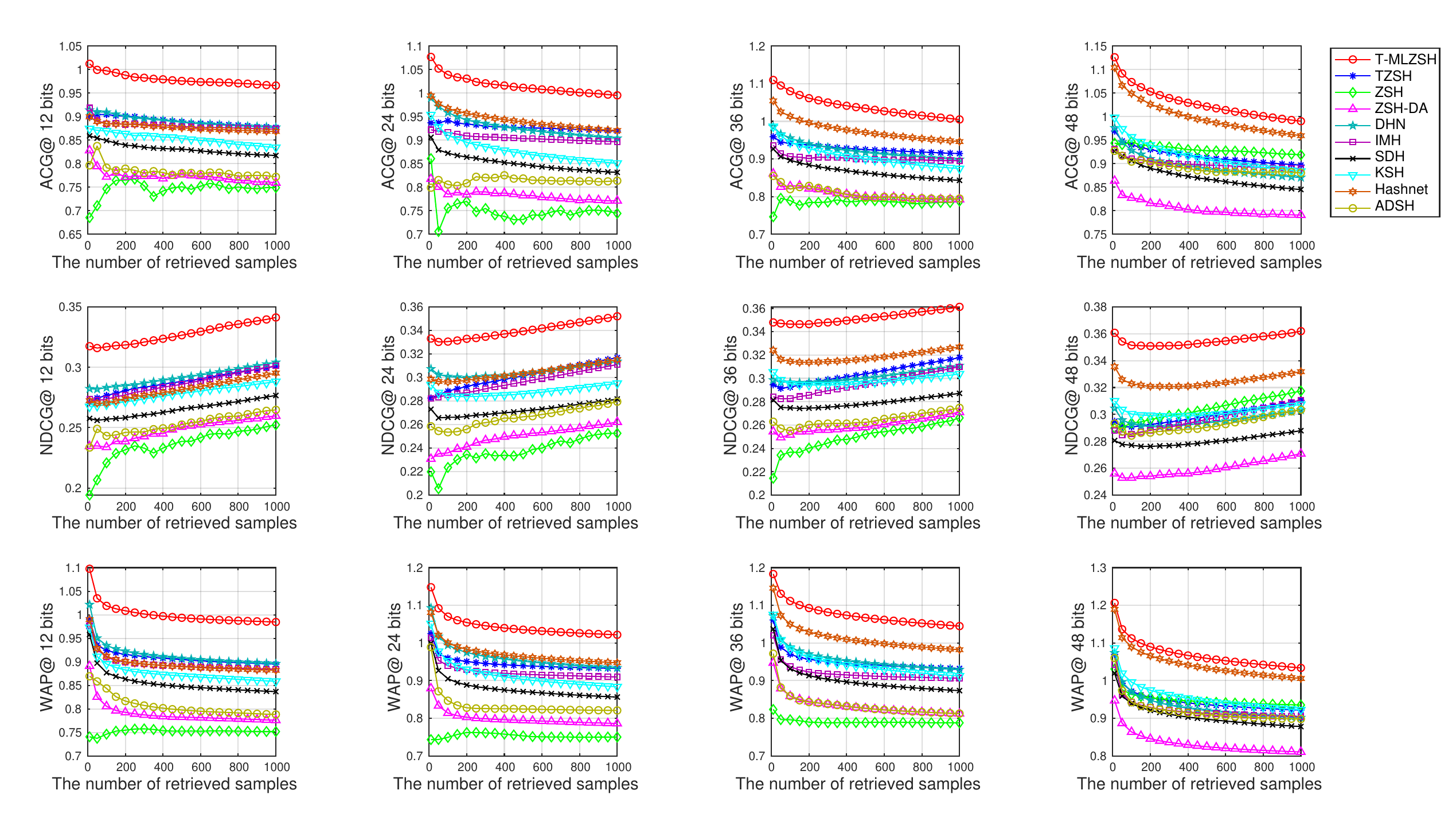}\\
	\caption{Performance comparison on VOC2012$\rightarrow$NUS-WIDE. The NUS-WIDE dataset is unseen. From top to bottom, there are ACG, NDCG, and WAP w.r.t. different top returned
		samples with hash codes of 12, 24, 36, and 48 bits, respectively.}
	\label{fig:nus}
\end{figure*}

\begin{figure*}[!t]
	\centering
	\includegraphics[width=0.98\linewidth]{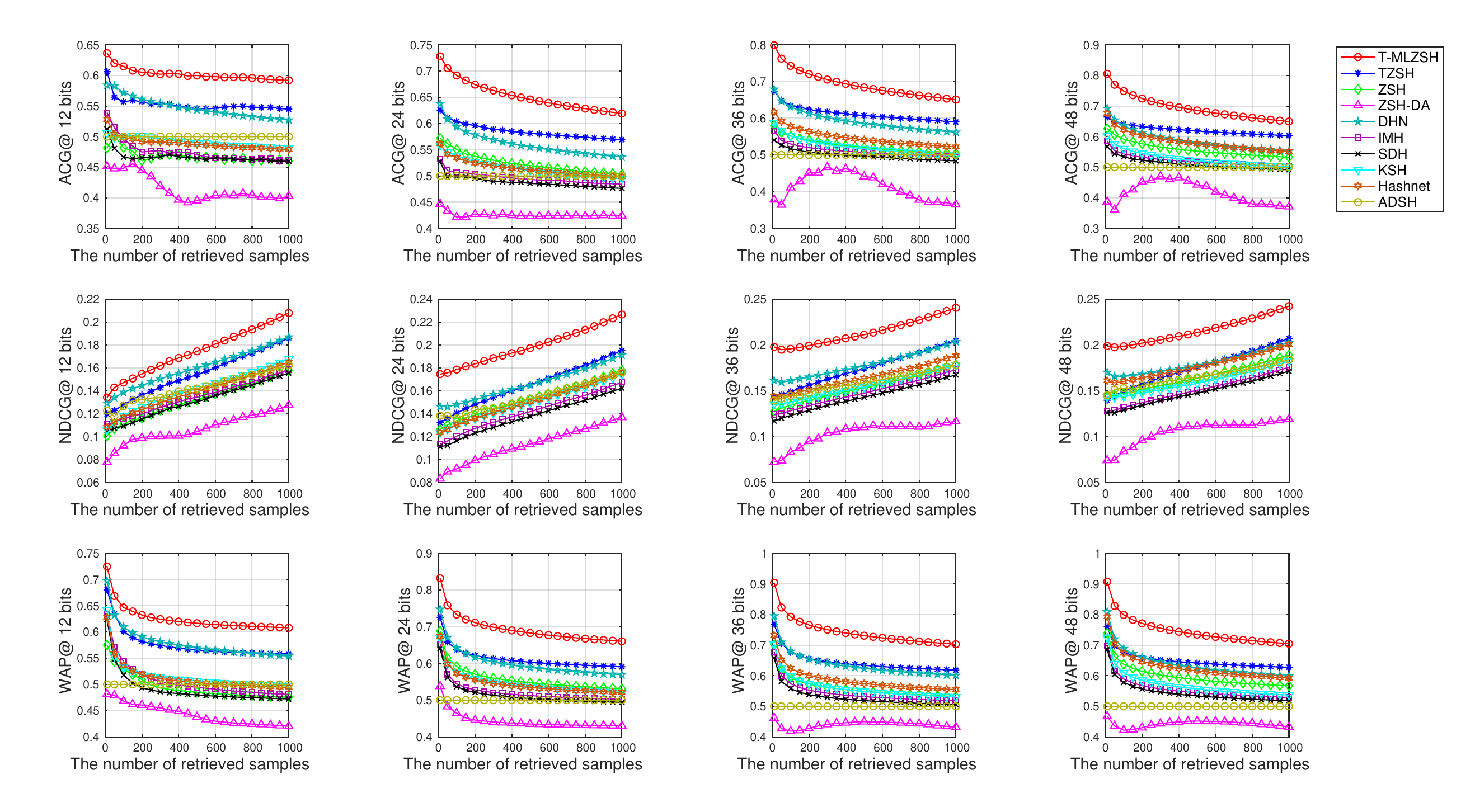}\\
	\caption{ Performance comparison on NUS-WIDE$\rightarrow$COCO. The COCO dataset is unseen. From top to bottom, there are ACG, NDCG, and WAP w.r.t. different top returned
		samples with hash codes of 12, 24, 36, and 48 bits, respectively.}
	\label{fig:coco}
\end{figure*}

\begin{figure*}[!t]
	\centering
	\includegraphics[width=0.98\linewidth]{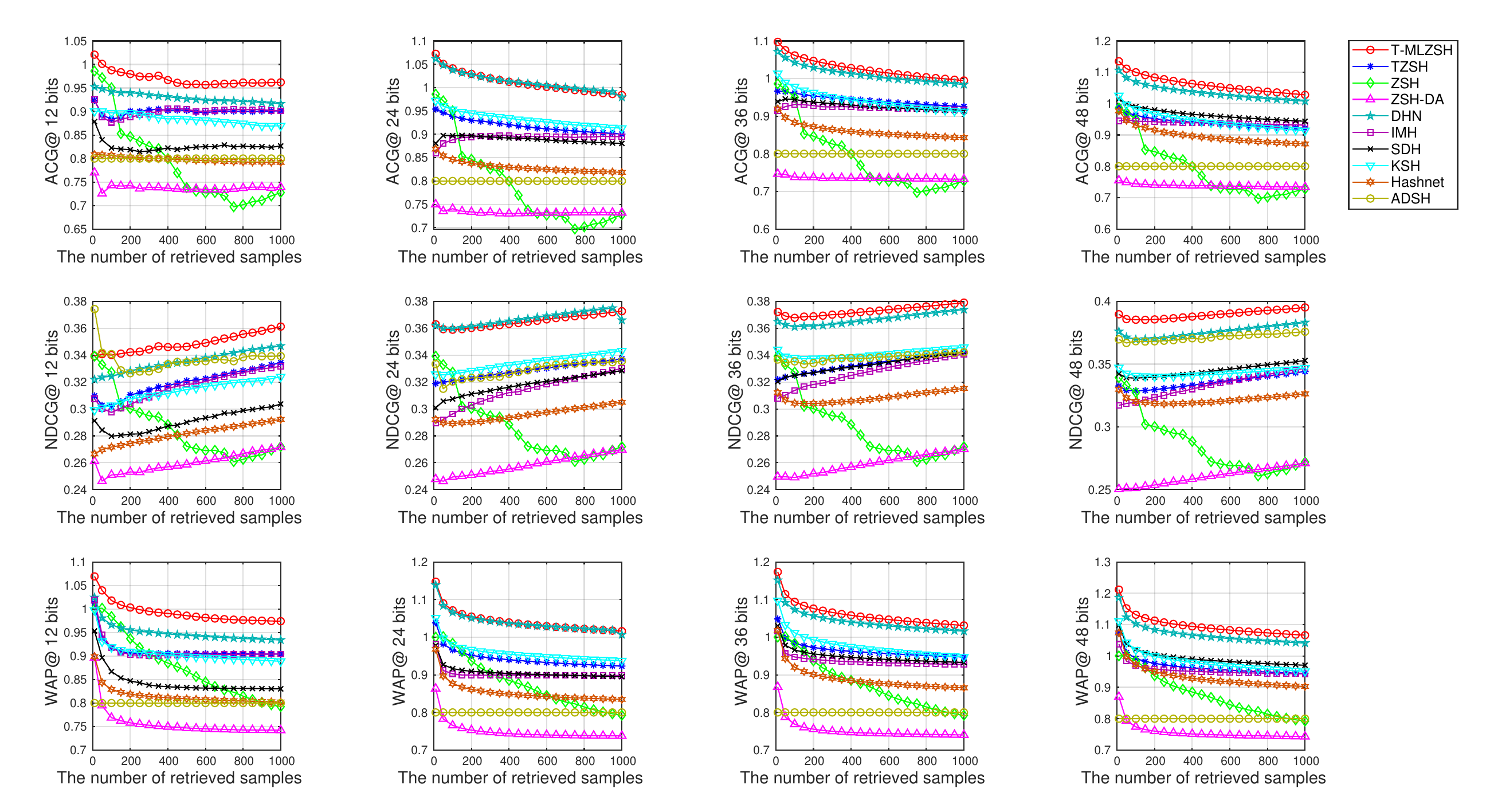}\\
	\caption{Performance comparison on COCO$\rightarrow$NUS-WIDE. The NUS-WIDE dataset is unseen. From top to bottom, there are ACG, NDCG, and WAP w.r.t. different top returned
		samples with hash codes of 12, 24, 36, and 48 bits, respectively.}
	\label{fig:nus17}
\end{figure*}

\begin{figure*}[!t]
	\centering
	\includegraphics[width=0.98\linewidth]{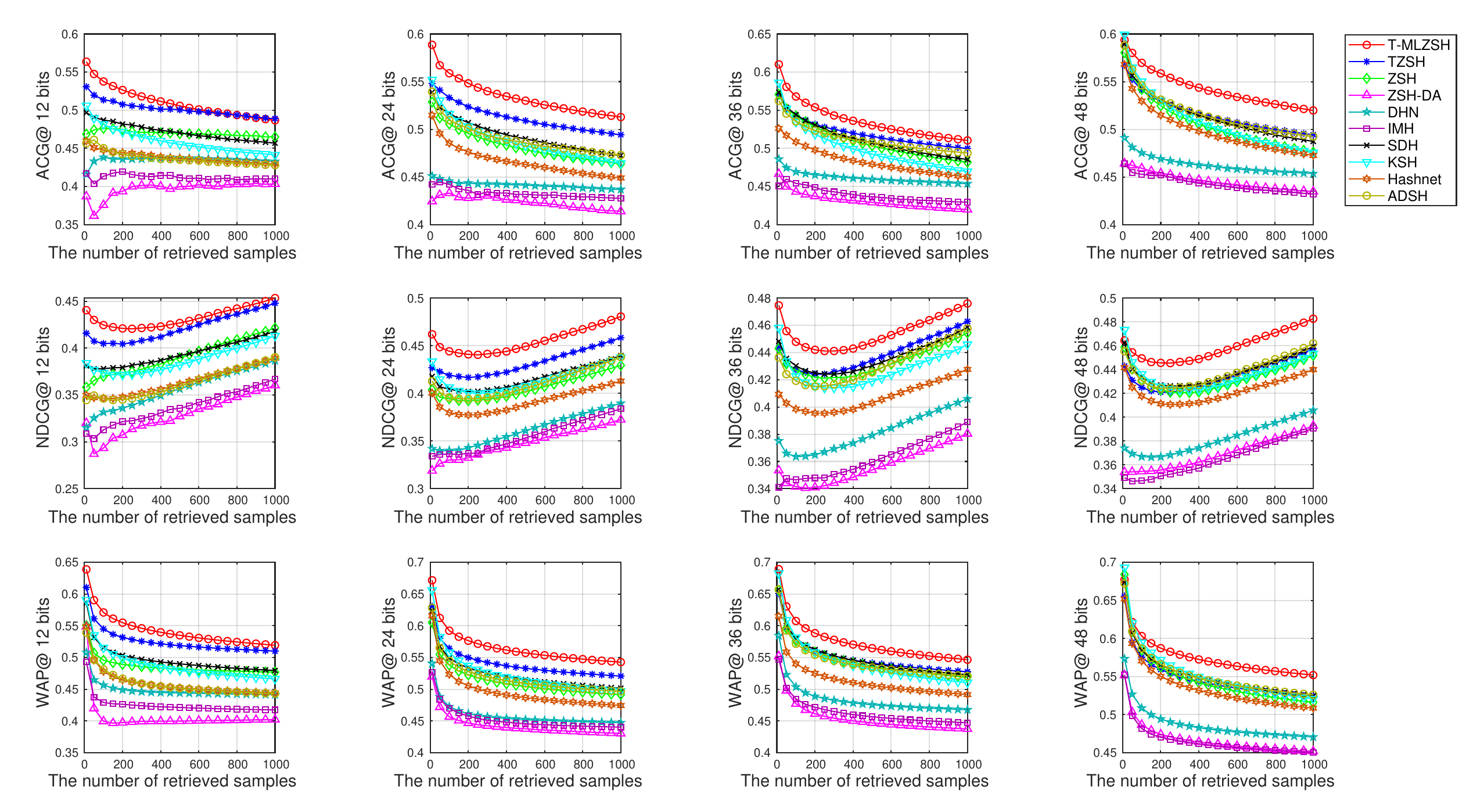}\\
	\caption{Performance comparison on COCO$\rightarrow$VOC2012. The VOC2012 dataset is unseen. From top to bottom, there are ACG, NDCG, and WAP w.r.t. different top returned samples with hash codes of 12, 24, 36, and 48 bits, respectively.}
	\label{fig:pascal2}
\end{figure*}

\begin{figure*}[!t]
	\centering
	\includegraphics[width=0.98\linewidth]{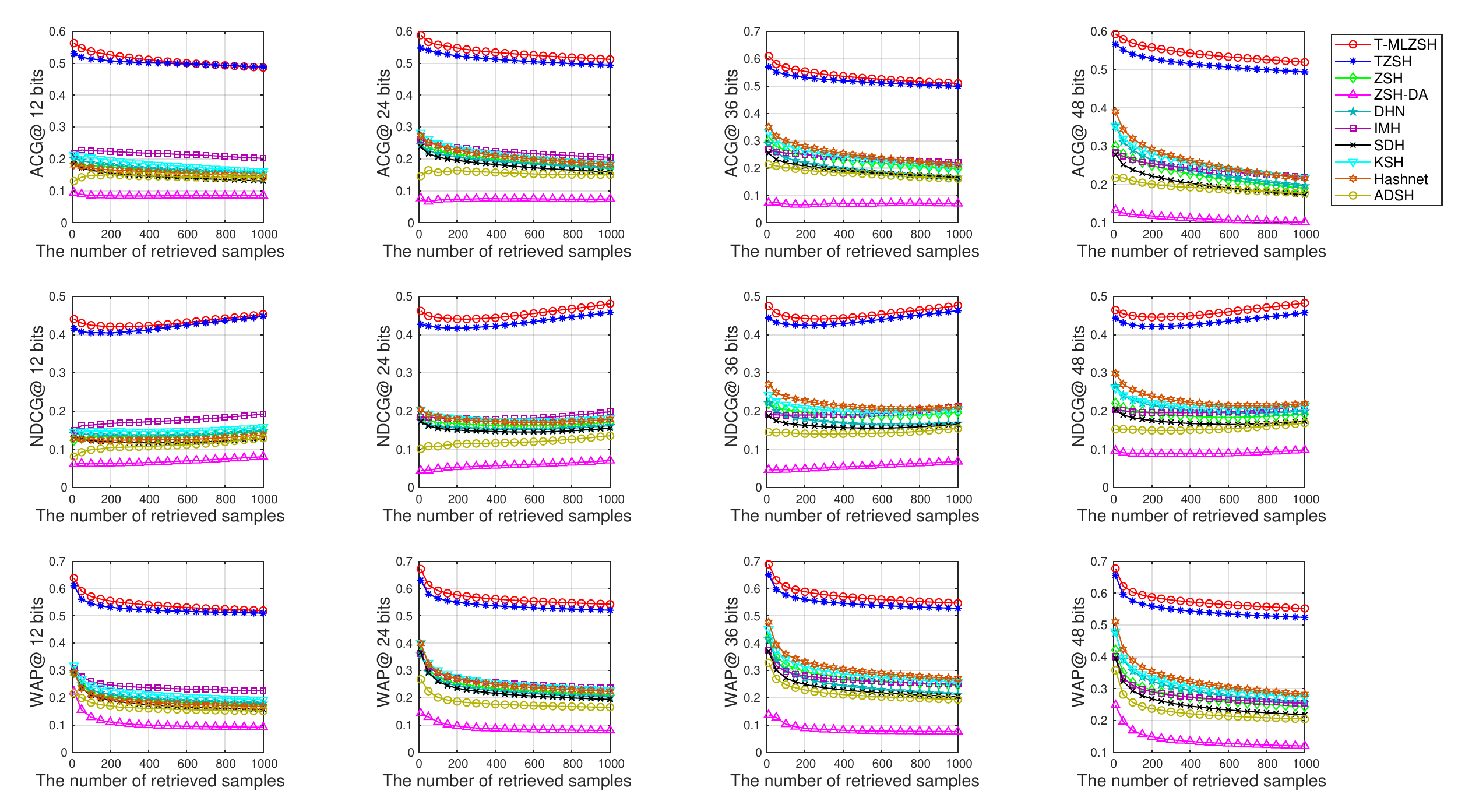}\\
	\caption{Performance comparison on VOC2012$\rightarrow$COCO. The COCO dataset is unseen. From top to bottom, there are ACG, NDCG, and WAP w.r.t. different top returned samples with hash codes of 12, 24, 36, and 48 bits, respectively.}\label{fig:coco60}
\end{figure*}

Figure~\ref{fig:pascal}-\ref{fig:pascal2} display more detailed results on ACG, NDCG, and WAP of different numbers of top returned images under different scenarios. Generally, the NDCG curves show a trend of first decreasing and then increasing while ACG and WAP decrease with the increase of retrieved samples. The more samples retrieved, the more low-quality retrieval results having fewer shared class labels will appear in the rear. This will lead to the trends that tend to be gentle. Since most of the compared methods are not specific for zero-shot learning, the results of unseen data may be uncertain. The curve of ZSH-DA is unsmooth. The possible reason is that, while using ZSH-DA for multi-label unseen images, the complex semantics are too hard to be modeled by learning a one-to-one semantic representation, which leads to unstable performances.

\begin{table*}[htbp]
    \caption{Performance under different backbone CNN blocks. The values are computed on the top-1000 retrieved images.}
	\centering
	\resizebox{\textwidth}{!}{
		\begin{tabular}{|c|c|c|c|c|c|c|c|c|c|c|c|c|c|c|c|c|}
			\hlinew{1.2pt}
			& \multicolumn{4}{c|}{MAP}                                                                                                                                          & \multicolumn{4}{c|}{NDCG}                                                                                                                                                                                         & \multicolumn{4}{c|}{ACG}                                                                                                                                          & \multicolumn{4}{c|}{WAP}                                                                                                                                                                                                                  \\ \cline{2-17}
			\multirow{-2}{*}{Metrics} & {12b}  & {24b}  &{36b}  &{48b}  & {12b}  & {24b}  &{36b}  &{48b} & {12b}  & {24b}  &{36b}  &{48b} & {12b}  & {24b}  &{36b}  &{48b}                                   \\ \hlinew{1.2pt}
			& \multicolumn{16}{c|}{NUS-WIDE-\textgreater{}VOC2012}  \\ \hline
			{T-MLZSH-AlexNet}                  &  {0.4808}         &{0.4884 }         &{0.4894 }          &  {0.5037  }       &  {0.4689  }      &  {0.4827  }        &  {0.4855 }        &  {0.4990 }      &  {0.5298}          & {0.5484 }         & {0.5545}         & {0.5716 }        & {0.5298 }       &  {0.5484}          & {0.5545 }        &  {0.5716 }        \\ \hline
			{T-MLZSH-VGG16}       &  {0.5340 }        &  {0.5346 }         &  {0.5843 }        &  {0.6149   }      &   {0.6363}         &  {0.6270 }          &   {0.6520}         &   {0.6782 }         &  {0.5299 }        & {0.5124  }        &  {0.5546 }         & { 0.5798 }         &   {0.5504  }       &   {0.5521    }      &   {0.6065  }        &   {0.6379 }        \\ \hline
			{T-MLZSH-ResNet50}
			&  {\textbf{0.5711} }&  {\textbf{0.6054}}& { \textbf{0.6518}} & { \textbf{0.6541 }} &  { \textbf{0.6471 }} & {  \textbf{0.6820 }} &   {\textbf{0.6915 }} &   {\textbf{0.7377 }} & { \textbf{0.5505 }} &  {\textbf{0.5761 }} &  {\textbf{0.5850 }} & { \textbf{0.6171 }} &   {\textbf{0.5902 }} &   {\textbf{0.6274 }} &  { \textbf{0.6739 }} &   {\textbf{0.6791 }} \\ \hline
			& \multicolumn{16}{c|}{VOC2012-\textgreater{}NUS-WIDE}\\ \hline
			{T-MLZSH-AlexNet}                   & { 0.6106 }         &  {0.6131}          &  {0.6149}          &  {0.6200}          &   {0.3412 }         &   {0.3520}          &   {0.3613  }        &  { 0.3619}          &  {0.9660 }       &  {1.0207}          &  {1.0448}          &  {1.0338}          &   {0.9845}          &   {1.0217}          &   {1.0448}          &   {1.0338}          \\ \hline
			{T-MLZSH-VGG16}                     &{  0.6049}          &  {0.6503}          &  {0.6635}          & { 0.6799}          &   {0.3280}          &   {0.3670}          &   {0.3657}          &   {0.3885}          &  {0.9175}          & { 1.0104}          &  {1.0238}          &  {1.0587}          &   {0.9315}          &  { 1.0361}          &   {1.0597}          &  { 1.1040}          \\ \hline
			{T-MLZSH-ResNet50}                  &  {\textbf{0.6542 }} & { \textbf{0.6715 }} &  {\textbf{0.6957 }} & { \textbf{0.6977 }} &  { \textbf{0.3677 }} &   {\textbf{0.3836 }} &  { \textbf{0.4056 }} &   {\textbf{0.4037 }} &  {\textbf{1.0267 }} &  {\textbf{1.0599 }} &  {\textbf{1.1146 }} & { \textbf{1.0882 }} &   {\textbf{1.0485 }} &  { \textbf{1.0858 }} &  { \textbf{1.1373 }} &   {\textbf{1.1282 }} \\ \hline
			& \multicolumn{16}{c|}{NUS-WIDE-\textgreater{}COCO} \\ \hline
			{T-MLZSH-AlexNet}         & {0.4724}    & {0.4941}    &{ 0.5291}      & {0.5124}   &  { 0.2097}          &   {0.2296}          &   {0.2405}          &   {0.2423}          &  {0.5921}          &  {0.6191}          & { 0.6508}          &  {0.7046}          &  { 0.6080}          &   {0.6604}          &   {0.6508}          &   {0.7046}          \\ \hline
			{T-MLZSH-VGG16}       & {0.4970}   & {\textbf{0.5705}}  & {0.6031} & {0.6075}   &  { 0.3014}          &  { \textbf{0.3683} } &   {0.3978}          &   {0.3935}          &  {0.6504}          & { \textbf{0.6724 }} &  {0.6889}          &  {0.7093}          &  { 0.6374}          &   {\textbf{0.6609 }} &   {0.6657}          &   {0.6700}          \\ \hline
			{T-MLZSH-ResNet50}         &{ \textbf{0.4994}}   & {0.5519}                                 &{ \textbf{0.6030}}                        &{ \textbf{0.6209}}                        &  { \textbf{0.3022 }} &   {0.3639}         &   {\textbf{0.3983 }} &   {\textbf{0.4112 }} &  {\textbf{0.6543 }} &  {0.6698}          & { \textbf{0.6934 }} &  {\textbf{0.7102 }} &   {\textbf{0.6309} } &   {0.6532 }         &  { \textbf{0.6694} } & {  \textbf{0.6741 }} \\ \hline
			
			& \multicolumn{16}{c|}{COCO-\textgreater{}NUS-WIDE}  \\ \hline
			{T-MLZSH-AlexNet }    &{0.6374} & {0.6425}                                 &{ 0.6510}   &{ 0.6693}                                 &{   0.3614}          &   {0.3728 }         &   {0.3791}          &   {0.3953}          &  {0.9621}          &{ 0.9014}          &{ 0.9948}          &{ 1.0280}          &{  0.9741}          &{  1.0162}          &{  1.0314 }         &{  1.0665}          \\ \hline
			{T-MLZSH-VGG16}                     &{0.6333}                                &{0.6686}                                 &{0.6709}                                 &{0.6858 }                                &{  0.4380}          &{  0.4808}          &{  0.4798}          &{  0.4985}          &{ 0.9718}          &{ 0.9796}          &{ 1.0297}          &{ 1.1145}          &{  0.9932}          &{  1.0345}          &{  1.1136 }         &{  1.1203}          \\ \hline
			{T-MLZSH-ResNet50}                  &{\textbf{0.6588}}                        &{\textbf{0.6843}}                        &{\textbf{0.6883}}                       &{\textbf{0.6914}}                        &{  \textbf{0.4713} } &{  \textbf{0.4925 }} &{  \textbf{0.5044 }} &{  \textbf{0.5020} } &{ \textbf{0.9821 }} &{ \textbf{0.9902} } &{ \textbf{1.1342 }} &{ \textbf{1.1201 }} &{  \textbf{1.0023 }} &{  \textbf{1.0962} } &{  \textbf{1.1361 }} &{  \textbf{1.1903 }} \\ \hline
			&\multicolumn{16}{c|}{VOC2012-\textgreater{}COCO} \\ \hline
			{T-MLZSH-AlexNet}  &{0.2465}     &{0.3149}  &{0.3670 }   &{0.3740 }   &{  0.2336 }         &{  0.2857}          &{  0.3304}          &{  0.3406}          &{ 0.2443}          &{ 0.2983}          &{ 0.3399}          &{ 0.3515}          &{  0.2642}          &{  0.3455}          &{  0.4050}          &{  0.4097}          \\ \hline
			{T-MLZSH-VGG16}      &{0.3707}     &{0.4183}  &{0.4757}     &{0.4901}    &{  0.3631 }         &{  0.3890}          &{  0.4430}          &{  0.4651}          &{ 0.4033}          &{ 0.4504}          &{ 0.5151}          &{ 0.4295}          &{  0.3968 }         &{  0.4234}          &{  0.4596}          &{  0.4775}          \\ \hline
			{T-MLZSH-ResNet50}                &{\textbf{0.4161}}                        &{\textbf{0.4695}}                        &{\textbf{0.5184}}                        &{\textbf{0.5320}}                        &{  \textbf{0.3992} } &{  \textbf{0.4372 }} &{  \textbf{0.4766} } &{  \textbf{0.4921 }} &{ \textbf{0.4558 }} &{ \textbf{0.5052} } &{ \textbf{0.5673 }} &{ \textbf{0.5804 }} &{  \textbf{0.4033} } &{  \textbf{0.4504} } &{  \textbf{0.5151 }} &{  \textbf{0.5295} } \\ \hline
			
			&\multicolumn{16}{c|}{COCO-\textgreater{}VOC1012} \\ \hline
			{T-MLZSH-AlexNet}                   &{ 0.5047}          &{ 0.5074}          &{ 0.5088}          &{ 0.5283}          &{  0.4502}          &{  0.4567}          &{  0.4665}          &{  0.4860}          &{ 0.4868}          &{ 0.4791}          &{ 0.4930}          &{ 0.5134}          &{  0.5193}          &{  0.5244}          &{  0.5338}          &{  0.5555}          \\ \hline
			{T-MLZSH-VGG16}                     &{ 0.5583}          &{ 0.6538}          &{ 0.6652 }         &{ \textbf{0.6844 }} &{  0.5131 }         &{  0.6084}          &{  0.6100}          &{  \textbf{0.6314 }} &{ 0.5761}          &{ \textbf{0.6804 }} &{ \textbf{0.6905}} &{ \textbf{0.7146 }} &{  0.5506 }         &{  0.6149}          &{  0.6224}          &{  0.6298}          \\ \hline
			{T-MLZSH-ResNet50}                  &{ \textbf{0.6402} } &{ \textbf{0.6672} } &{ \textbf{0.6796 }} &{ 0.6761}          &{  \textbf{0.6056 }} &{  \textbf{0.6148 }} &{  \textbf{0.6330 }} &{  0.6272}          &{ \textbf{0.6210 }} &{ 0.6223}          &{ 0.6327}          &{ 0.6274}          &{  \textbf{0.6614 }} &{  \textbf{0.6939 }} &   {\textbf{0.7071 }} &   {\textbf{0.7027 }} \\ \hlinew{1.2pt}
	\end{tabular}}
	\label{table:four}
\end{table*}

\subsubsection{Comparison with different backbone CNN blocks}
To justify the versatility of the proposed deep hashing framework, We replace the backbone CNN with VGG16 and ResNet50, both of which achieve more accurate results than AlexNet on the ImageNet competition. We denote these two modifications as `T-MLZSH-VGG16' and `T-MLZSH-ResNet50', respectively. The results are shown in Table~\ref{table:four}. We can see from the results that, with more powerful backbone CNN blocks,
T-MLZSH generally achieves higher performance on all these metrics. It simply indicates a good transfer capability and versatility of the proposed deep hashing framework.

\subsection{Functional Analysis}
\subsubsection{Influence of the categories of datasets}~ In the above two groups of experiments, part of them used NUS-WIDE as the unseen dataset and the results are shown in Table~\ref{table:one} and Table~\ref{table:two}. From the left of the tables, we notice that using COCO as seen dataset can achieve a better MAP result, which has an improvement of about 2.6\%, 2.9\%, 3.6\%, and 4.9\% in average MAP with different hash bits respectively. These two groups of experiments have the same target domain and the only difference is the source domain. We guess that the possible reason leading to different MAPs is the difference of categories. COCO dataset is more finely divided and more semantic information can be used which make the network much stronger.

\begin{figure}[!t]
	\centering
	\centering
	\subfigure[\scriptsize{NUS-WIDE $\rightarrow$ COCO}]{
		\includegraphics[width=0.48\linewidth]{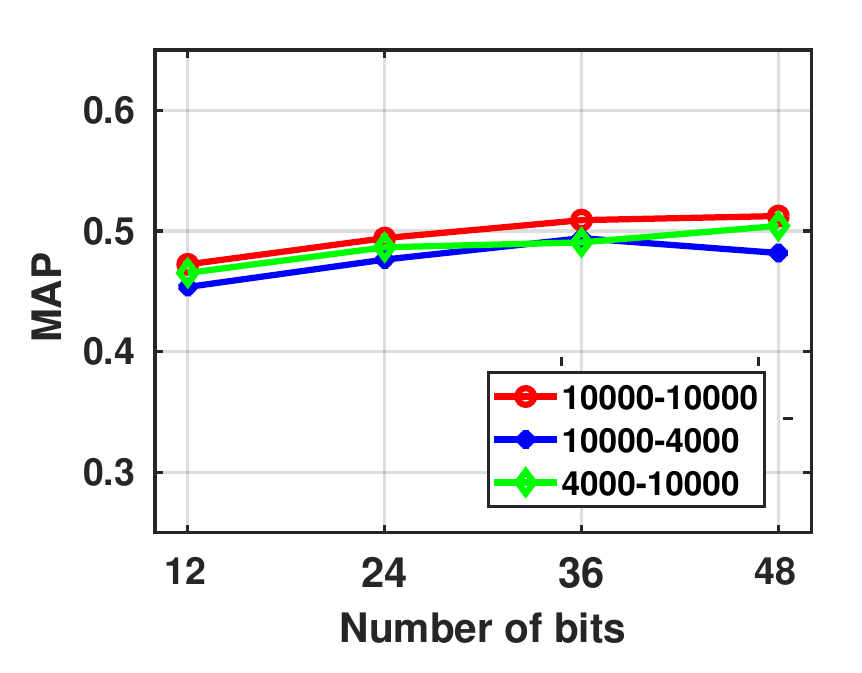}
	}\hspace{-3mm}
	\subfigure[\scriptsize{COCO $\rightarrow$ NUS-WIDE}]{
		\includegraphics[width=0.48\linewidth]{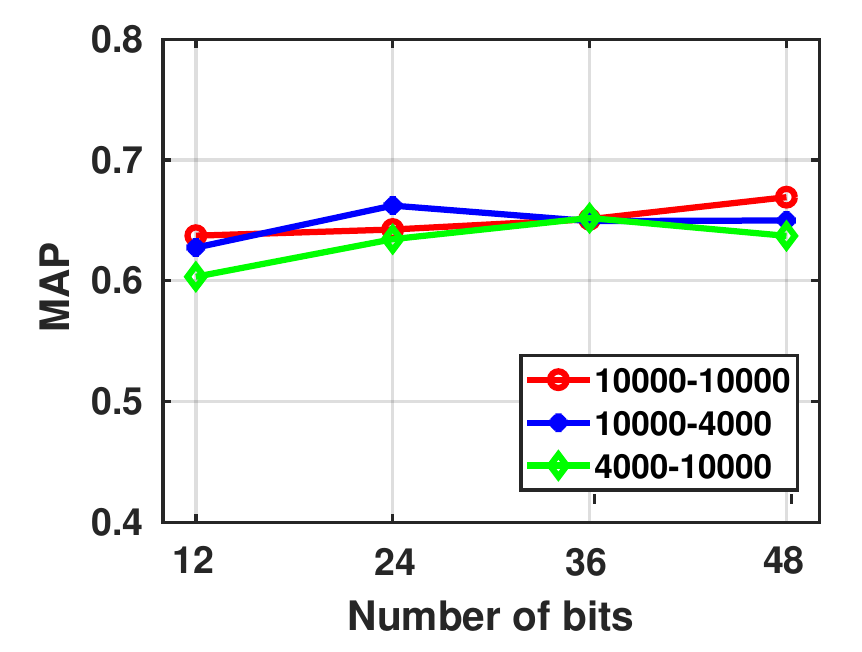}
	}\vspace{-3mm}
	\caption{Performance obtained by using different amounts of training data. The numbers indicate the training data in the form of `source - target'.}
	\label{fig:num_compare}
\end{figure}

\subsubsection{Influence of the size of datasets}~ Moreover, we explore the influence of the sizes of the source and target datasets by using different amounts of `seen' and `unseen' images to train the model. Three orders of magnitude are considered. Exactly, the number of images from source dataset and target dataset are 10,000 and 10,000, 10,000 and 4,000, 4,000 and 10,000, respectively. The results are shown in Fig.~\ref{fig:num_compare}. It can be seen that there is a slight difference in the use of different orders of magnitude, but the performance is stable on the whole. It manifests that the proposed model has certain stability even if the number of images from two domains used for training varies from each other.

\begin{figure}[!t]
	\centering
	\centering
	\subfigure[\scriptsize{NUS-WIDE $\rightarrow$ VOC2012}]{
		\includegraphics[width=0.45\linewidth]{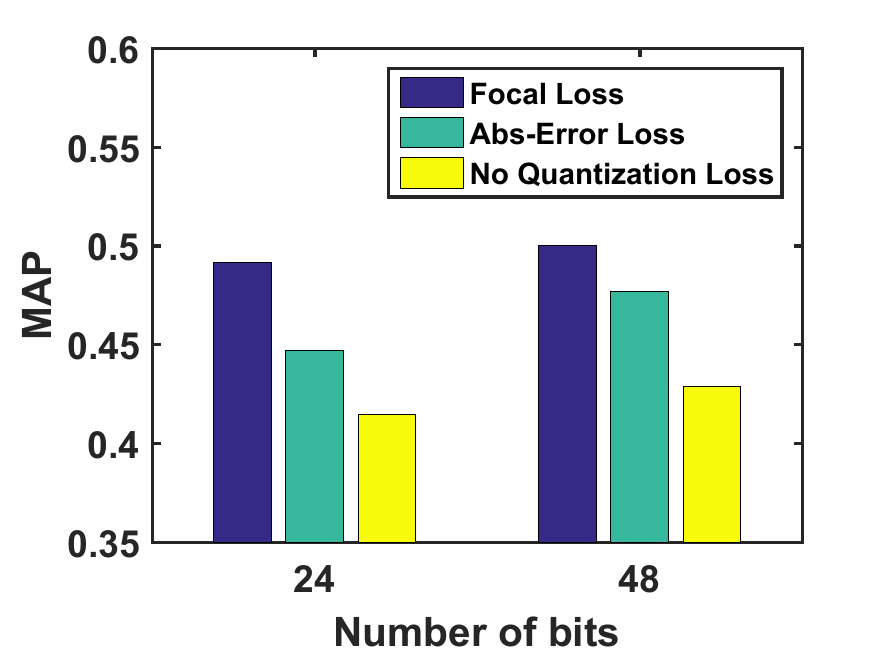}
	}
	\subfigure[\scriptsize{VOC2012 $\rightarrow$ NUS-WIDE}]{
		\includegraphics[width=0.45\linewidth]{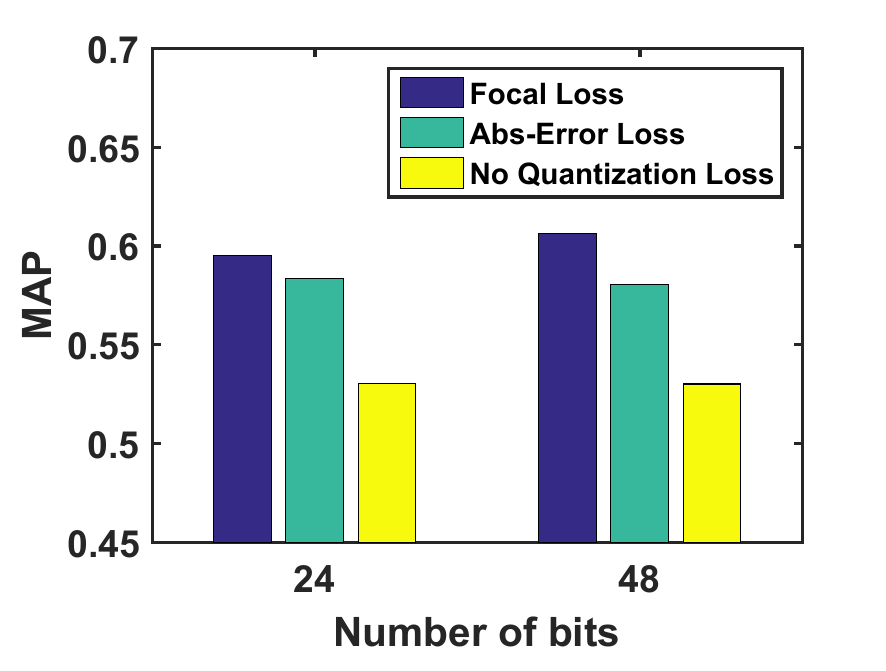}
	}\vspace{-3mm}
	\caption{Performance obtained by different quantization losses.}
	\label{fig:loss_compare}
\end{figure}

\subsubsection{Necessary of quantization loss}~We also explore the effectiveness of the proposed quantization loss. We compare the proposed method with its variant versions: one adopts the widely used absolute error loss that measures the Euclidean distance between continuous outputs and discrete codes directly, and the other does not use quantization loss. The results are presented in Fig.~\ref{fig:loss_compare}. It can be seen that, without quantization loss, there is a rapid degradation of the performance. The difference in the evaluation index of the MAP is about 0.5\%, which illustrates the importance of using quantization loss in deep hashing learning. We can also see that, applying quantified losses has greatly improved the results, the proposed focal quantization loss has a much more advanced performance among all the proposed architecture.

\begin{figure}[!t]
	\centering
	\centering
	\subfigure[\scriptsize{Predicted label}]{
		\includegraphics[width=0.45\linewidth]{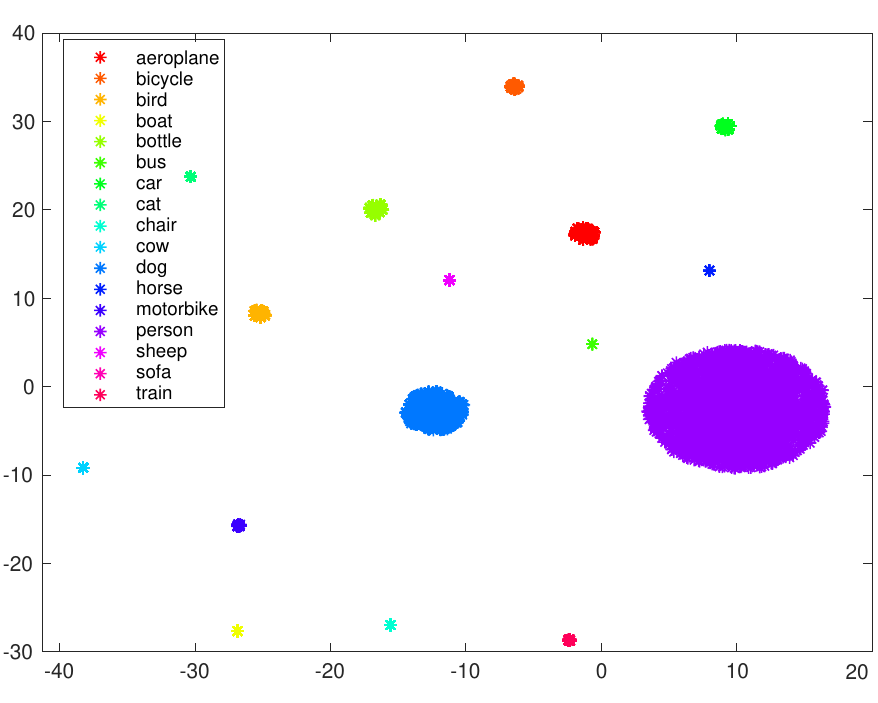}
	}
	\subfigure[\scriptsize{True label}]{
		\includegraphics[width=0.45\linewidth]{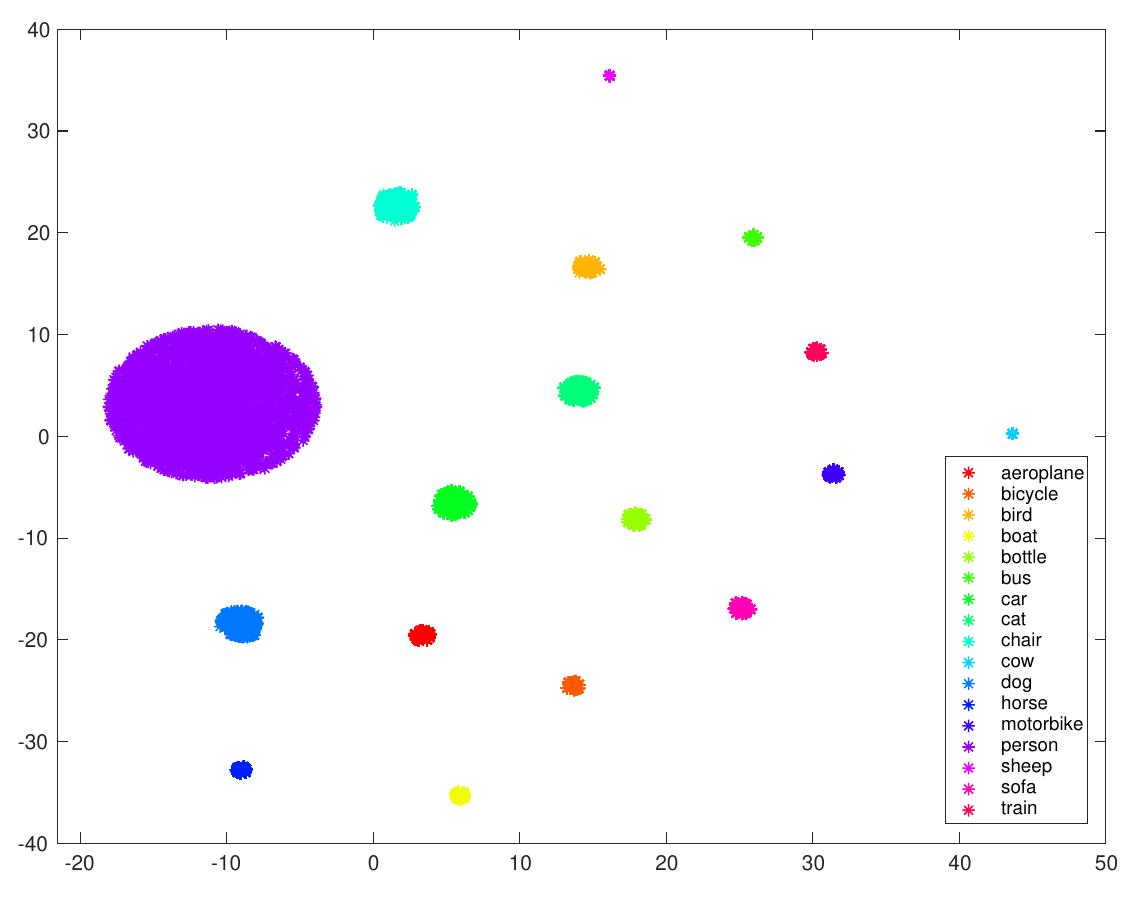}
	}\vspace{-3mm}
	\caption{Visualization of the predicted labels and the true labels using t-SNE.}
	\label{fig:pred label}
\end{figure}

\subsubsection{Effectiveness of the visual-semantic bridge}
The visual-semantic bridge is built to predict labels for unseen data. It helps the images belonging to the same category be with higher similarity by transferring the knowledge from the semantic representations to visual features. Since the labels of unseen images are unknown, the predicted labels will be represented as meaningless digital codes for the unseen images.

To validate the effectiveness of the visual-semantic bridge in linking the semantic representations and visual features, we use the proposed model to predict labels for the seen images and compare them to true labels. In the experiment, we use the model trained on the training set of  VOC2012 to predict the labels for 10,000 images from the database set. Since the average number of objects in each image from VOC2012 is 1.301, we predict one label for each image. The distribution of labels is visualized by t-SNE, as shown in Fig.~\ref{fig:pred label}. Comparing Fig.~\ref{fig:pred label}(a) and (b), we can find that the overall distributions are similar for the true label and the predicted label. However, because one image may contain multiple labels but is predicted with only one, the area of each predicted label category is observed to be a little smaller than that of the true label category. Meanwhile, we count the correctness of the predicted labels. The predicted labels for 8,283 images fall within their true labels. These experimental results show that the proposed visual-semantic bridge has a high performance.

\begin{figure*}[t!]
	\centering
	\includegraphics[width=0.98\linewidth]{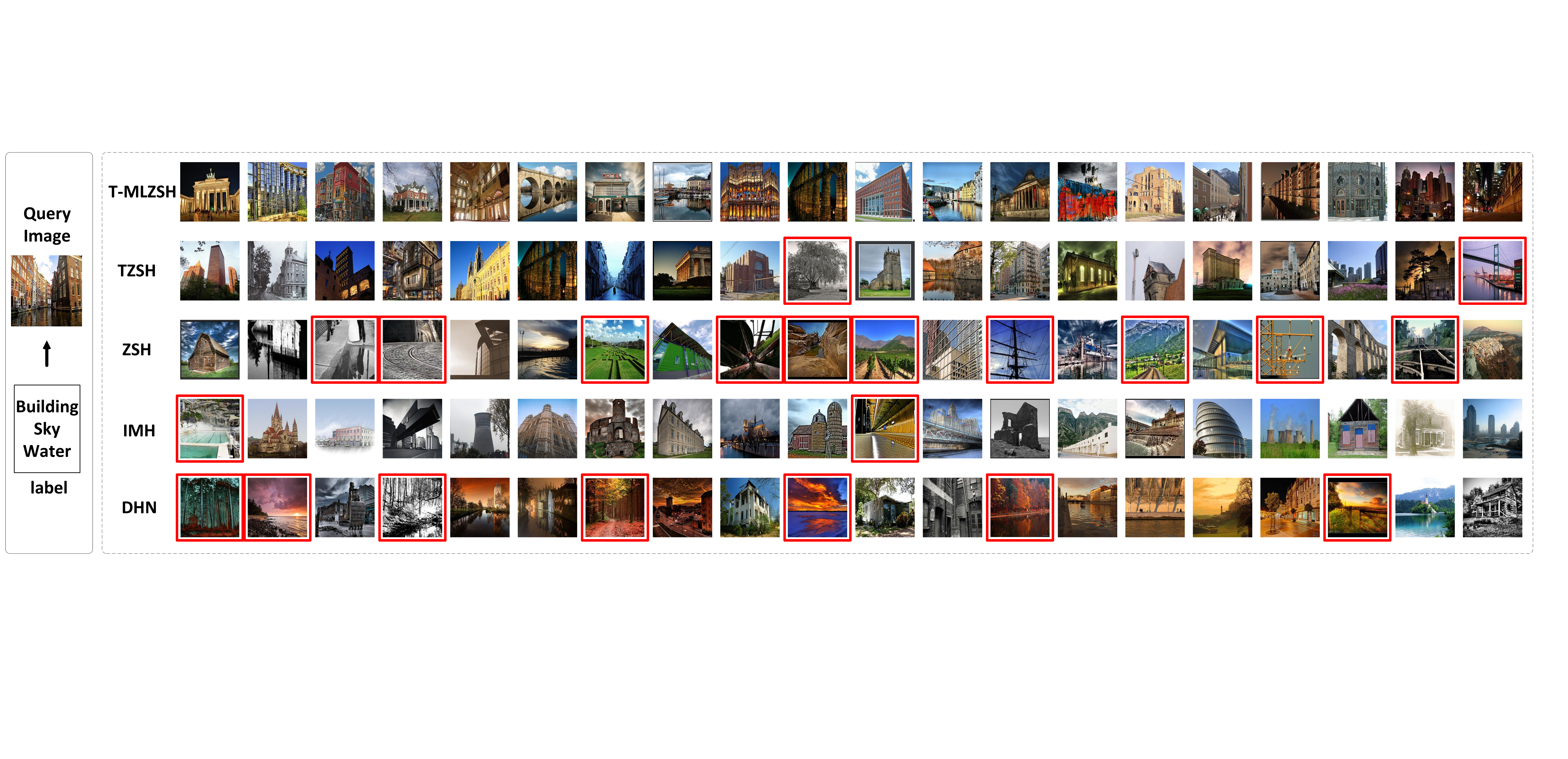}\\
	\caption{Top-20 images retrieved by the proposed method and the comparison methods using the Hamming ranking on 48-bit hash codes.}
	\label{fig:examples}
\end{figure*}

\subsubsection{Top retrieval results}
Figure~\ref{fig:examples} shows some retrieval samples of some typical hashing methods according to the ascending Hamming ranking. The query image contains three semantic labels, \textit{i.e.}, building, sky, and water, with the main content of a building. We mark the mismatched image with the red box from the perspective of human vision. The retrieval results of T-MLZSH are better visual plausible while focusing on the main object of the query image, while other compared methods may return some mismatched results like the forest, or return some images related to the less important part of the query image with higher ranking orders.

\subsubsection{Running efficiency} We do the image retrieval by returning the top 1,000 similar images from 16,900 images based on the trained models. The running time includes calculating the hash codes and calculating the hamming distances. In our experiments, when calculating the 36-bit hash codes of 16,901 images, it takes 17.4616s, 48.2418s and 34.6509s for T-MLZSH-AlexNet, T-MLZSH-VGG16 and T-MLZSH-ResNet50, respectively. That is, it will take about 1.03ms, 2.85ms, and 2.05ms for the three methods to calculate the hash codes of a new image. The hash codes are computed on an NVIDIA TITAN Xp GPU. The GPU memory usage is about 2.5GB, 4.3GB, and 1.2GB for the three models in the retrieval test.  For calculating the hamming distance of 36-bit hash codes, the running time on the 16,900 images is 0.006742s, which is about 4$\times$10$^{-7}$s for one calculation. The hamming distance is computed by a 2.0GHz core of an Intel(R) Xeon(R) E5-2620 CPU.

\section{Conclusion}\label{sec:conc}
In this paper, a novel transductive zero-shot hashing method was proposed for multi-label image retrieval. It utilized the instance-concept coherence to construct a bridge for connecting the seen and unseen labels. Based on these connections, several categories with the highest relatedness scores were selected as the predicted labels for target data. Then, hashing learning was performed on both the source data and target data in a supervised manner. Experimental results on three widely-used multi-label datasets showed that the proposed method outperformed state-of-the-art methods with a significant margin. Moreover, the superiority of the proposed focal loss was verified by ablation studies, the effectiveness of the visual-semantic bridge was demonstrated through feature visualization, and the high performance on image retrieval was illustrated with visual comparisons.

\section*{Acknowledgements}\label{sec:conc}
The authors would like to thank Dr. Song~Bai from the University of Oxford for helpful suggestions.

\bibliographystyle{IEEEtran}
\bibliography{refs}

\vspace{12mm}

{\footnotesize
\noindent \textbf{Qin Zou} received the B.E. degree in information engineering in 2004 and the Ph.D. degree in computer vision in 2012, from Wuhan University, China. From  2010  to  2011, he was a visiting PhD student at the Computer Vision Lab, University of South Carolina, USA. Currently, he is an associate professor with the School of Computer Science, Wuhan University. He is a co-recipient of the National Technology Invention Award of China in 2015. His research activities involve computer vision, pattern recognition, and machine learning. He is a senior member of the IEEE and a member of the ACM.

\vspace{10mm}

\noindent \textbf{Ling Cao} received the B.S. degree in computer science from Yunnan University in 2018, and is now working towards her M.S. degree in computer application at the School of Computer Science, Wuhan University, China. Her research interests include deep learning and image/video retrieval.

\vspace{10mm}

\noindent \textbf{Zheng Zhang} received the B.S. degree in computer science and M.S. degree in computer application from the School of Computer Science, Wuhan University, in 2016 and 2019, respectively.
He won the first prize in `China Undergraduate Contest in Internet of Things' in 2015. His research interests include deep learning and its application in image classification and retrieval.

\vspace{10mm}

\noindent \textbf{Long Chen} received the B.Sc. degree in communication engineering and the Ph.D. degree in signal and information processing from Wuhan University, Wuhan, China, in 2007 and in 2013, respectively. From October 2010 to November 2012, he was co-trained PhD Student at National University of Singapore. He is currently an Associate Professor with the School of Data and Computer Science, Sun Yat-sen University, Guangzhou, China. His areas of interest include deep learning and cognitive perception.

\vspace{10mm}

\noindent \textbf{Song Wang} received the PhD degree in electrical and computer
engineering from the University of Illinois at Urbana-Champaign
(UIUC) in 2002. From 1998 to 2002, he also worked as a research
assistant in the Image Formation and Processing Group at the Beckman
Institute of UIUC. In 2002, he joined the Department of Computer
Science and Engineering at the University of South Carolina, where
he is currently a professor. His research interests include computer
vision, medical image processing, and machine learning. He is
currently serving as an Associate Editor of IEEE Transactions on Pattern Analysis and Machine Intelligence and an Associate Editor of Pattern Recognition
Letters. He is a senior member of the IEEE and the IEEE Computer
Society.
}

\end{document}